\documentclass[11pt]{article}
\PassOptionsToPackage{table}{xcolor}
\PassOptionsToPackage{hyperfootnotes=false}{hyperref}
\usepackage[preprint]{acl}
\usepackage{times}
\usepackage{latexsym}
\usepackage[T1]{fontenc}
\usepackage[utf8]{inputenc}
\usepackage{microtype}
\usepackage{inconsolata}
\usepackage{graphicx}
\usepackage{booktabs}
\usepackage{multirow}
\usepackage{makecell}
\usepackage{amsmath}
\usepackage{amssymb}
\usepackage{subcaption}
\usepackage{algorithm}
\usepackage{algpseudocode}
\graphicspath{{figures/}}

\title{MultiAlloc: Budget-Aware Precision Allocation for KV Cache Quantization}
\author{
  \textbf{Yangchengyu Zhou}\textsuperscript{1}\thanks{Equal contribution.}
  \quad \textbf{Zhiping Wu}\textsuperscript{1,*}
  \quad \textbf{Runchang Jiang}\textsuperscript{2}
  \quad \textbf{Dongdong Ren}\textsuperscript{3} \\
  \textbf{Youfang Han}\textsuperscript{4}
  \quad \textbf{Wei Zhao}\textsuperscript{4}
  \quad \textbf{Dongqi Cai}\textsuperscript{1}
  \quad \textbf{Wenbin Li}\textsuperscript{1}\thanks{Corresponding author: \texttt{liwenbin@nju.edu.cn}.} \\[3pt]
  \textnormal{\small\textsuperscript{1}Nanjing University} \\
  \textnormal{\small\textsuperscript{2}The Hong Kong University of Science and Technology} \\
  \textnormal{\small\textsuperscript{3}Geely\quad\textsuperscript{4}Goertek} \\[3pt]
  \textnormal{\small\texttt{zhouyangchengyu@smail.nju.edu.cn}\quad\texttt{zhipingwu@smail.nju.edu.cn}} \\
  \textnormal{\small\texttt{rjiangaw@connect.ust.hk}\quad\texttt{Dongdong.Ren2@geely.com}} \\
  \textnormal{\small\texttt{fred.hanyf@goertek.com}\quad\texttt{charles.zhaow@goertek.com}} \\
  \textnormal{\small\texttt{cdq@nju.edu.cn}\quad\texttt{liwenbin@nju.edu.cn}} \\[3pt]
  \textnormal{\small\textbf{Code:} \url{https://github.com/RL-MIND/MultiAlloc}}
}
\hypersetup{pdfauthor={Yangchengyu Zhou, Zhiping Wu, Runchang Jiang, Dongdong Ren, Youfang Han, Wei Zhao, Dongqi Cai, Wenbin Li}}

\hypersetup{
  pdftitle={MultiAlloc: Budget-Aware Precision Allocation for KV Cache Quantization},
  pdfsubject={KV cache quantization and budget-aware precision allocation},
  pdfkeywords={MultiAlloc, KV cache, quantization, large language models}
}

\begin{document}
\maketitle
\begin{abstract}
The Key-Value (KV) cache constitutes a dominant source of memory consumption during large language model (LLM) inference, particularly in long-context and complex reasoning scenarios.
Existing KV cache quantization methods have explored different compression granularities, including layer-wise precision adjustment and token-level information preservation.
In practical deployment, however, the key challenge is not merely compression, but how to allocate limited quantization resources among heterogeneous precision preservation strategies under a fixed KV cache memory budget.
We propose \emph{MultiAlloc}, a profiling-guided precision allocation framework for KV cache quantization that adaptively distributes limited quantization resources between layer-wise precision enhancement and token-wise preservation.
MultiAlloc employs lightweight offline profiling to characterize layer sensitivity and token importance, enabling adaptive precision allocation under a fixed memory constraint.
Guided by the profiled memory--accuracy trade-off, it automatically derives an effective precision allocation strategy without online allocation search, and applies the resulting policy during inference.
Extensive experiments on reasoning and long-context benchmarks demonstrate that MultiAlloc consistently improves over layer-wise, token-wise, and recent multi-granularity KV cache quantization methods under matched memory budgets across diverse models, including Mistral, Llama, DeepSeek-Qwen, and QwQ.
\end{abstract}

\section{Introduction}

\begin{figure}[t]
    \centering
    \includegraphics[width=1\columnwidth]{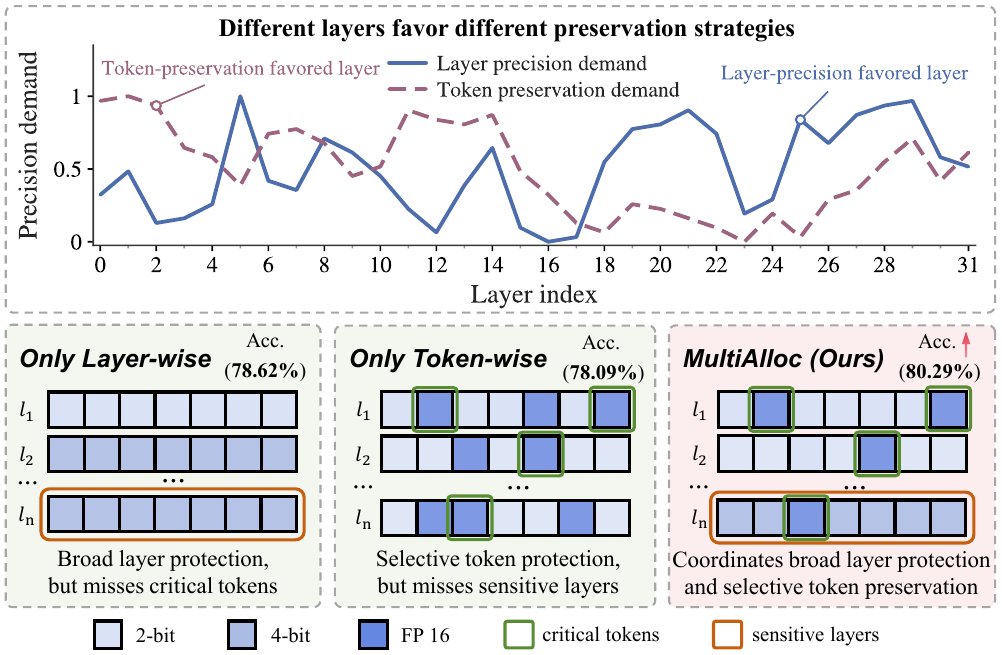}
    \caption{
    Motivation for budget-aware precision allocation under a fixed KV cache budget.
    Different layers exhibit heterogeneous demands for layer precision enhancement
    and token preservation. Existing layer-wise and token-wise methods optimize
    individual preservation strategies, while \textit{MultiAlloc} profiles these
    demands and allocates the shared memory budget across preservation strategies.
}
    \label{fig:motivation}
    \vspace{-20pt}
\end{figure}

As large language models (LLMs) become increasingly integral to long-context reasoning, complex question answering, and multi-step planning, their growing inference-time memory footprint has emerged as a critical bottleneck to practical deployment~\cite{xiao2024efficient,
10.1145/3600006.3613165,liu2026pmkvq}. 
In long-sequence generation, the
Key--Value (KV) cache constitutes a dominant source of GPU memory consumption,
making efficient KV cache compression a central challenge for scalable and
cost-effective LLM inference~\cite{zhang-etal-2026-mixkvq,
wang-etal-2026-hqekv}.

Recent studies on KV cache quantization have developed increasingly targeted strategies to reduce memory overhead while preserving model accuracy.
KIVI~\cite{liu2024kivi} introduces asymmetric low-bit quantization by applying channel-level quantization to keys and token-level quantization to values, substantially reducing KV cache memory consumption.
Subsequent token-oriented methods further improve this direction, with ZipCache~\cite{NEURIPS2024_7e57131f} identifying salient tokens and OTT~\cite{su-etal-2025-accurate} tracing and preserving outlier tokens at higher precision to reduce token quantization errors.
However, these methods largely overlook the heterogeneous sensitivity of different layers.
In contrast, KVTuner~\cite{li2025kvtuner} and PM-KVQ~\cite{liu2026pmkvq} allocate mixed precision according to layer or block sensitivity, thereby reducing quantization errors caused by layer heterogeneity.
Nevertheless, they generally apply uniform token treatment within each layer and do not explicitly preserve token-level heterogeneity.
Recent multi-granularity methods, such as KVmix~\cite{li2026kvmix}, further explore the joint adaptation of layer-wise precision and token-level preservation by incorporating layer importance and pivotal context selection.
However, their token-level adaptation mainly focuses on context selection rather than precision allocation based on heterogeneous token importance.

Motivated by these observations, we formulate KV cache quantization as a precision allocation problem under a fixed memory budget.
We propose \textit{MultiAlloc}, a profiling-guided precision allocation framework that characterizes heterogeneous precision demands across layers and tokens.
Specifically, \textit{MultiAlloc} employs layer-wise sensitivity profiling to estimate the precision requirements of different layers, while token-wise importance profiling identifies tokens that benefit from high-precision preservation.
Based on these complementary profiling signals, \textit{MultiAlloc} adaptively determines the allocation of limited quantization resources between layer-wise precision enhancement and token-wise high-precision preservation.
All allocation decisions are derived offline and applied as a static inference-time policy, eliminating online allocation search during inference.

We conduct extensive experiments on reasoning and long-context benchmarks, including GSM8K, GPQA, and LongBench-v2, using models ranging from 7B to 32B parameters, including Llama-3-8B-Instruct~\cite{grattafiori2024llama}, Mistral-7B-Instruct-v0.3~\cite{jiang2023mistral}, DeepSeek-R1-Distill-Qwen-7B/14B~\cite{deepseekai2025deepseekr1incentivizingreasoningcapability}, and QwQ-32B~\cite{qwq32b}.
The results demonstrate that \textit{MultiAlloc} consistently outperforms representative layer-wise, token-wise, and importance-based KV cache quantization methods under matched KV cache memory budgets, with robust improvements across different model scales and both reasoning and long-context tasks.

Our contributions are summarized as follows:
\begin{itemize}

\item We reveal heterogeneous precision demands across layers and tokens in KV cache quantization, motivating precision allocation under a fixed KV cache memory budget.

\item We propose \textit{MultiAlloc}, a profiling-guided precision allocation framework that models layer-wise sensitivity and token importance to adaptively allocate limited quantization resources between layer precision enhancement and token preservation.

\item Extensive experiments across diverse models and reasoning and long-context benchmarks demonstrate that \textit{MultiAlloc} consistently outperforms representative layer-wise, token-wise, and importance-based KV cache quantization methods under matched KV cache memory budgets.

\end{itemize}

\section{Related Work}
\label{sec:related_work}

\textbf{KV Cache Quantization and Precision Allocation.}
Recent KV cache quantization methods explore different strategies to reduce quantization errors while preserving inference quality.
KIVI~\cite{liu2024kivi} introduces asymmetric low-bit quantization by applying channel-level quantization to keys and token-level quantization to values.
Token-aware approaches, such as ZipCache~\cite{NEURIPS2024_7e57131f} and OTT~\cite{su-etal-2025-accurate}, preserve salient tokens at higher precision to mitigate token quantization errors.
Layer-wise methods, including KVTuner~\cite{li2025kvtuner} and PM-KVQ~\cite{liu2026pmkvq}, allocate mixed precision according to layer sensitivity.
KVmix~\cite{li2026kvmix} further explores layer- and token-level adaptation, while KV-COBRA~\cite{ha2026kvcobra} studies budget-aware allocation of compression configurations under fixed memory constraints.
MoE-nD~\cite{sun2026moend} extends KV compression to multiple dimensions by jointly considering quantization and token reduction.
However, existing approaches mainly optimize individual compression strategies or configurations, without explicitly modeling heterogeneous precision demands across granularities or allocating a shared budget between different precision preservation strategies.

\textbf{KV Cache Memory Management.}
Beyond quantization, another line of work focuses on managing KV cache memory usage during LLM inference.
PagedAttention~\cite{10.1145/3600006.3613165} improves memory efficiency through virtual memory paging, while subsequent approaches such as H2O~\cite{NEURIPS2023_6ceefa7b}, StreamingLLM~\cite{xiao2024efficient}, SqueezeAttention~\cite{wang2024squeezeattention}, XKV~\cite{li2024xkvpersonalizedkvcache}, and KVPR~\cite{jiang-etal-2025-kvpr} reduce memory pressure through token eviction, retention, or offloading strategies.
These methods primarily optimize which KV entries to retain or where to store them, whereas \textit{MultiAlloc} focuses on allocating limited quantization resources among different precision preservation strategies within the KV cache representation.

\textbf{Differentiation of Our Work.}
Existing KV cache quantization methods have demonstrated the effectiveness of preserving important layers or tokens at higher precision.
However, they mainly focus on identifying important components or adapting individual precision strategies, without explicitly modeling heterogeneous precision demands across different granularities.
In contrast, \textit{MultiAlloc} introduces a profiling-guided precision allocation framework that determines how limited quantization resources should be distributed between layer-wise precision enhancement and token-wise high-precision preservation under a fixed KV cache memory budget.

\section{Preliminary}
\label{sec:background}

\subsection{Generative Inference and KV Cache}
\label{subsec:kv_cache_background}

Transformer-based Large Language Models (LLMs) generate text autoregressively in two phases: \emph{prefilling} and \emph{decoding}.
Given a prompt sequence of length $T_{in}$, \emph{prefilling} processes all tokens in parallel to generate the initial Key-Value (KV) cache, while \emph{decoding} generates new tokens token-by-token.
To avoid recomputing historical representations, the Key and Value vectors of past tokens are cached in GPU memory.
For an $L$-layer Transformer, let $\mathbf{h}_{l,t} \in \mathbb{R}^{d}$ denote the hidden state of the $t$-th token at layer $l$. The query, key, and value vectors are computed via linear projections:
\begin{equation}\small
\mathbf{q}_{l,t}=\mathbf{W}^{Q}_{l}\mathbf{h}_{l,t},\quad
\mathbf{k}_{l,t}=\mathbf{W}^{K}_{l}\mathbf{h}_{l,t},\quad
\mathbf{v}_{l,t}=\mathbf{W}^{V}_{l}\mathbf{h}_{l,t}.
\end{equation}
where $\mathbf{W}^Q_l, \mathbf{W}^K_l, \mathbf{W}^V_l \in \mathbb{R}^{d \times d}$ are learnable weight matrices.
At decoding step $t > T_{in}$, the query $\mathbf{q}_{l,t}$ attends to all cached keys and values. The KV cache is updated by concatenating the new vectors along the sequence dimension:
\begin{equation}\small
\begin{split}
    \mathbf{K}_{l, \le t} &= \text{Concat}(\mathbf{K}_{l, <t}, \mathbf{k}_{l,t}), \\
    \mathbf{V}_{l, \le t} &= \text{Concat}(\mathbf{V}_{l, <t}, \mathbf{v}_{l,t}).
\end{split}
\end{equation}
The attention output $\mathbf{o}_{l,t}$ is:
\begin{equation}
    \mathbf{o}_{l,t} = \text{softmax}\left( \frac{\mathbf{q}_{l,t} \mathbf{K}_{l, \le t}^\top}{\sqrt{d}} \right) \mathbf{V}_{l, \le t}.
\end{equation}
As the sequence length $T$ grows, the KV cache footprint scales linearly as $O(L \cdot T \cdot d)$, often exceeding model-weight memory in long-context scenarios and becoming a primary inference bottleneck.

\subsection{KV Cache Quantization}
\label{subsec:quantization_background}

To alleviate this bottleneck, Post-Training Quantization (PTQ) compresses the KV cache into low-bit representations.
For a tensor $\mathbf{X}$ (representing either $\mathbf{K}$ or $\mathbf{V}$), uniform affine quantization maps floating-point values to $b$-bit integers:
\begin{equation}
    \bar{\mathbf{X}} = \text{clamp}\left(\left\lfloor \frac{\mathbf{X}}{s} + z \right\rceil, 0, 2^b-1\right),
\end{equation}
where $s$ and $z$ denote the scale factor and zero-point, respectively. The dequantized approximation is reconstructed as $\hat{\mathbf{X}} = (\bar{\mathbf{X}} - z) \cdot s$.
The quantization error $\mathcal{E} = \|\mathbf{X} - \hat{\mathbf{X}}\|_F^2$ stems primarily from rounding and outlier clamping.
Conventional quantization approaches often apply uniform bit-widths across layers and tokens, overlooking the heterogeneous precision requirements within KV cache representations.

However, this overlooks the heterogeneous sensitivity of KV cache representations: (1) certain layers are more robust to quantization noise and can tolerate lower bit-widths, while (2) specific tokens contribute disproportionately to quantization errors due to their distinct value distributions.

\section{Method}

\begin{figure*}[t]
    \centering
    \includegraphics[width=\textwidth]{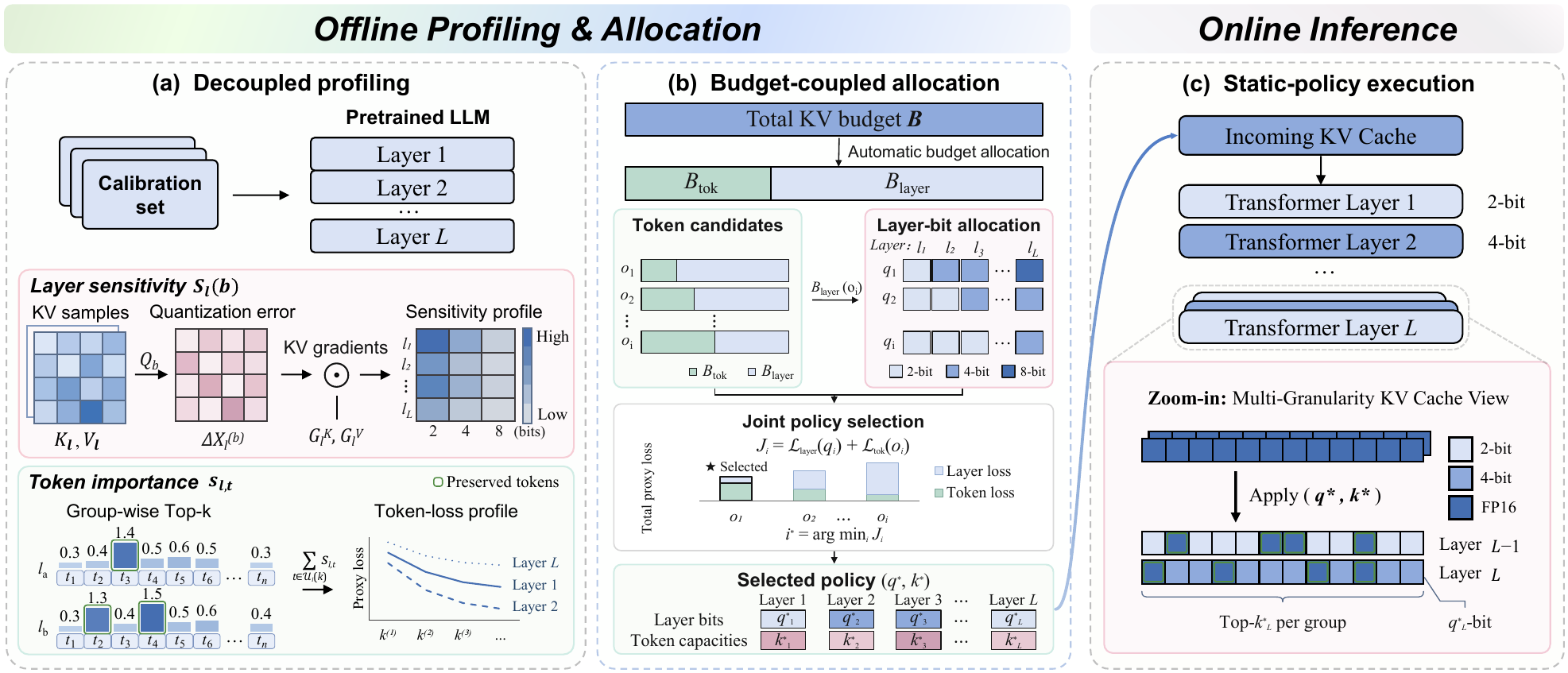}
    \caption{
    \textbf{Overview of \textit{MultiAlloc}}.
    An offline profiling stage estimates layer-wise precision demands and
    token-wise preservation demands, followed by a budget-coupled precision
    allocation module that derives the static policy
    $(\mathbf{q}^{\star},\mathbf{k}^{\star})$ under a fixed KV cache memory
    budget $B$.
    The policy is then directly applied during online inference.
}
    \label{fig:framework}
    \vspace{-10pt}
\end{figure*}

We propose \textit{MultiAlloc}, a profiling-guided precision allocation
framework for KV cache quantization.
As illustrated in Figure~\ref{fig:framework}, \textit{MultiAlloc} consists of
three stages: decoupled precision demand profiling, budget-coupled precision
allocation, and static policy execution.

\subsection{Problem Formulation and Overview}
\label{subsec:problem}

We formulate KV cache quantization as a precision allocation problem under
a fixed KV cache memory budget.
Given a Transformer model with $L$ layers, we aim to determine a static
quantization policy that allocates limited quantization resources between
layer-wise precision enhancement and token-wise high-precision preservation.

Specifically, let 
$\mathbf{q}=\{q_l\}_{l=1}^{L}$ denote the layer-wise bit-width
configuration, where $q_l$ represents the quantization precision of the KV
cache at layer $l$.
Meanwhile, let
$\mathbf{k}=\{k_l\}_{l=1}^{L}$ denote the token preservation configuration,
where $k_l$ represents the number of high-precision tokens preserved per
group at layer $l$.
The resulting policy is represented as
$\pi=(\mathbf{q},\mathbf{k})$.

Given a fixed KV cache memory budget $B$, the objective is to minimize the
quantization-induced degradation while satisfying the memory budget:
\begin{equation}
\min_{\mathbf{q},\mathbf{k}} \mathcal{L}(\mathbf{q},\mathbf{k})
\quad
\mathrm{s.t.}\quad
\mathcal{M}(\mathbf{q},\mathbf{k})\leq B .
\end{equation}
where $\mathcal{L}(\cdot)$ denotes the estimated degradation caused by KV
cache compression and $\mathcal{M}(\cdot)$ denotes the corresponding memory
consumption.

Directly optimizing this objective requires searching over a large number of
layer-wise precision and token preservation configurations.
To address this challenge, \textit{MultiAlloc} adopts an offline profiling and
allocation pipeline, as illustrated in Figure~\ref{fig:framework}.
First, the decoupled profiling stage estimates layer-wise precision demands
and token-wise preservation demands from calibration data.
Then, the budget-coupled precision allocation module evaluates candidate
policies and determines the allocation of quantization resources between
different preservation strategies.
Finally, the selected static policy is directly applied during online
inference without additional allocation search.

\subsection{Layer-wise Precision Demand Estimation}
\label{subsec:layer_demand}

Different Transformer layers exhibit different sensitivity to KV cache
quantization, resulting in heterogeneous requirements for quantization
precision.
To estimate layer-wise precision demands, \textit{MultiAlloc} profiles the
quantization-induced degradation of each layer under different bit-widths.

Let $\mathcal{Q}_{b}(\cdot)$ denote the quantization operator with bit-width
$b$.
For the KV cache of layer $l$, the quantization errors of keys and values are
defined as:
\begin{equation}\small
\Delta \mathbf{K}_{l}^{(b)}
=
\mathcal{Q}_{b}(\mathbf{K}_{l})-\mathbf{K}_{l},
\quad
\Delta \mathbf{V}_{l}^{(b)}
=
\mathcal{Q}_{b}(\mathbf{V}_{l})-\mathbf{V}_{l}.
\end{equation}

The impact of quantization errors on model output can be approximated by a
first-order Taylor expansion:
\begin{equation}\small
\Delta \mathcal{L}_{l}^{(b)}
\approx
\left\langle
\frac{\partial \mathcal{L}}{\partial \mathbf{K}_{l}},
\Delta \mathbf{K}_{l}^{(b)}
\right\rangle
+
\left\langle
\frac{\partial \mathcal{L}}{\partial \mathbf{V}_{l}},
\Delta \mathbf{V}_{l}^{(b)}
\right\rangle .
\end{equation}

In practice, we estimate the layer-wise degradation using the magnitude of
the first-order contribution over calibration samples:
\begin{equation}\small
S_l(b)
=
\mathbb{E}
\left[
\left\|
\Delta \mathbf{K}_{l}^{(b)}
\odot
\mathbf{G}_{l}^{K}
\right\|_1
+
\left\|
\Delta \mathbf{V}_{l}^{(b)}
\odot
\mathbf{G}_{l}^{V}
\right\|_1
\right],
\end{equation}
where
$\mathbf{G}_{l}^{K}=\frac{\partial\mathcal{L}}{\partial\mathbf{K}_{l}}$
and
$\mathbf{G}_{l}^{V}=\frac{\partial\mathcal{L}}{\partial\mathbf{V}_{l}}$.

The resulting profile
$\{S_l(b)\}_{l=1}^{L}$
captures the precision requirements of different layers under candidate
bit-widths.
These layer-wise precision demands are then provided to the
budget-coupled allocation module to determine the layer precision
configuration under the KV cache memory budget.

\subsection{Token-wise Preservation Demand Estimation}
\label{subsec:token_demand}

Besides layer-wise precision differences, KV cache quantization also exhibits
token-level heterogeneity, where a small subset of tokens can contribute
disproportionately to inference quality.
\textit{MultiAlloc} therefore estimates token-wise preservation demands to
determine how much high-precision storage should be allocated to different
layers.

For a token representation $\mathbf{K}_{l,t}$ at layer $l$, we measure its
importance using the activation magnitude:
\begin{equation}
s_{l,t}=\|\mathbf{K}_{l,t}\|_1,
\end{equation}
which provides a lightweight proxy for the contribution of token $t$ to
attention computation.

During inference, tokens are organized into groups of size $G$.
For each group, \textit{MultiAlloc} preserves the top-$k_l$ tokens with
higher precision according to their importance scores, while the remaining
tokens are quantized using the layer-wise assigned precision.
To estimate the preservation demand, we measure the importance mass lost after
preserving $k_l$ tokens in each group:
\begin{equation}
\mathcal{L}_{\mathrm{tok}}(k_l)
=
\sum_{t\in\mathcal{G}_{l}}s_{l,t}
-
\sum_{t\in\mathrm{TopK}(\mathcal{G}_{l},k_l)}s_{l,t},
\end{equation}
where $\mathcal{G}_{l}$ denotes a token group at layer $l$.

By evaluating different preservation capacities, we obtain a token-level
preservation profile:
\[
\{\mathcal{L}_{\mathrm{tok}}(k_l)\}_{l=1}^{L},
\]
which characterizes the precision demand of preserving additional tokens.
Together with the layer-wise precision profiles, these estimates provide
complementary signals for the subsequent budget-coupled precision allocation.

\subsection{Budget-Coupled Precision Allocation}
\label{subsec:allocation}

The layer-wise and token-wise profiling stages provide complementary
precision demand signals for allocation.
However, both precision enhancement strategies compete for the same KV cache
memory budget.
Therefore, \textit{MultiAlloc} performs budget-coupled precision allocation
to determine how limited quantization resources should be distributed between
layer precision enhancement and token preservation.

Given a KV cache memory budget $B$, we first construct a set of token
preservation candidates:
\begin{equation}
\mathcal{O}=\{o_i\}_{i=1}^{N},
\end{equation}
where each candidate $o_i$ specifies a layer-wise token preservation capacity
configuration:
\begin{equation}
o_i=\{k_{i,l}\}_{l=1}^{L}.
\end{equation}

For each candidate $o_i$, its token preservation memory cost
$M_{\mathrm{tok}}(o_i)$ and degradation
$\mathcal{L}_{\mathrm{tok}}(o_i)$ are estimated according to the token
preservation profiles.
The remaining memory budget is then assigned to layer-wise precision
allocation:
\begin{equation}
B_{\mathrm{layer}}(o_i)
=
B-M_{\mathrm{tok}}(o_i).
\end{equation}

Under the remaining layer budget, the optimal layer-wise bit-width
configuration is obtained by solving:
\begin{equation}
\begin{aligned}
q_i^{*}
=
\arg\min_{\mathbf{q}}\small
\quad &
\mathcal{L}_{\mathrm{layer}}(\mathbf{q})\\
\mathrm{s.t.}\quad&
M_{\mathrm{layer}}(\mathbf{q})
\leq B_{\mathrm{layer}}(o_i),
\end{aligned}
\end{equation}
where $\mathbf{q}=\{q_l\}_{l=1}^{L}$ denotes the layer-wise precision
configuration.

Each candidate policy is then ranked using a surrogate degradation score:
\begin{equation}
J_i
=
\mathcal{L}_{\mathrm{layer}}(q_i^{*})
+
\mathcal{L}_{\mathrm{tok}}(o_i).
\end{equation}

The final precision allocation policy is selected as:
\begin{equation}
i^{*}
=
\arg\min_i J_i,
\end{equation}
yielding the final policy
\[
\pi^{*}=(q^{*},k^{*}).
\]

By coupling token preservation decisions with layer-wise precision allocation,
\textit{MultiAlloc} avoids manually specified budget splits and adaptively
allocates quantization resources according to heterogeneous precision
demands.

\subsection{Static Policy Execution}
\label{subsec:execution}

After the offline allocation stage, \textit{MultiAlloc} obtains a static
precision policy $\pi^{*}$, which specifies the layer-wise precision
configuration and token preservation configuration.
During inference, the policy is directly applied to incoming KV caches
without additional allocation search.

Specifically, for each layer $l$, newly generated KV states are quantized
according to the assigned bit-width $q_l^{*}$.
Meanwhile, tokens are processed in groups according to the selected
preservation configuration $k^{*}$.
Within each group, the top-$k_l^{*}$ tokens with higher importance scores are
stored in high precision, while the remaining tokens are quantized using the
corresponding layer-wise precision.

The resulting KV cache combines quantized representations with preserved
high-precision tokens, enabling efficient long-context inference
under the fixed KV cache memory budget.

\section{Experiments}
\subsection{Experimental Setup}
\label{sec:exp_setup}

We evaluate reasoning and long-context performance on five instruction-tuned
models spanning architectures and scales:
Llama-3-8B-Instruct~\cite{grattafiori2024llama},
Mistral-7B-Instruct-v0.3~\cite{jiang2023mistral},
DeepSeek-R1-Distill-Qwen-7B and DeepSeek-R1-Distill-Qwen-14B~\cite{deepseekai2025deepseekr1incentivizingreasoningcapability},
and QwQ-32B~\cite{qwq32b}.
We report accuracy on GSM8K~\cite{cobbe2021training} and
GPQA~\cite{rein2023gpq}, together with macro-averaged accuracy across the six
task categories of LongBench-v2~\cite{bai2025longbench}.
The overall score averages the GPQA, GSM8K, and LongBench-v2 macro-average
scores.
Dataset statistics and evaluation details are provided in Appendix A.
All methods use identical prompts, tokenization, stopping criteria, and greedy
decoding. All experiments are conducted on NVIDIA A100 GPUs.

For each model, we specify a model-dependent KV cache budget and evaluate
all compressed methods under the same KV cache memory budget, with budgets
reported in Table~\ref{tab:main_results}.
We compare OTT~\cite{su-etal-2025-accurate}, a representative token-wise
preservation method, PM-KVQ~\cite{liu2026pmkvq}, a representative layer-wise
mixed-precision method, KVmix~\cite{li2026kvmix}, an importance-guided hybrid
KV cache quantization method, and our profiling-guided precision allocation
framework \textit{MultiAlloc}.
All methods are implemented within the same inference framework and follow a
unified KV cache memory-accounting protocol.

\subsection{Main Results under Fixed KV Budgets}
\label{sec:main_results}

Table~\ref{tab:main_results} reports the main results under fixed,
model-specific KV cache budgets. For each model, OTT, PM-KVQ, KVmix, and
\textit{MultiAlloc} are evaluated under the same memory budget, while FP16
serves as an uncompressed reference. Across all five models,
\textit{MultiAlloc} achieves the highest overall accuracy among compressed
methods, demonstrating the effectiveness of profiling-guided precision
allocation under constrained KV cache budgets.

\begin{table*}[t]
\centering
\small
\setlength{\tabcolsep}{7pt}
\renewcommand{\arraystretch}{1.15}

\begin{tabular}{llccccc}
\toprule
Model
& Method
& \makecell{KV Budget}
& GPQA
& GSM8K
& \makecell{LongBench-v2}
& Avg \\
\midrule

\multirow{5}{*}{Llama-3-8B-Instruct}

& FP16
& 1024 MB
& 32.83
& 80.89
& 24.79
& 46.17 \\

& OTT \textcolor{gray}{(ACL 2025)}
& 423 MB
& \textbf{32.83}
& 76.95
& 23.14
& 44.31 \\

& PM-KVQ \textcolor{gray}{(ICLR 2026)}
& 423 MB
& 30.30
& 79.98
& 24.56
& 44.95 \\

& KVmix \textcolor{gray}{(AAAI 2026)}
& 423 MB
& \textbf{32.83}
& 74.83
& \textbf{26.94}
& 44.87 \\

\cmidrule[0.35pt]{2-7}
\rowcolor{gray!8}
\cellcolor{white}
& \textbf{MultiAlloc (Ours)}
& \textbf{423 MB}
& \textbf{32.83}
& \textbf{80.36}
& 25.85
& \textbf{46.35} \\

\midrule

\multirow{5}{*}{Mistral-7B-Instruct}

& FP16
& 1024 MB
& 27.27
& 48.75
& 24.78
& 33.60 \\

& OTT \textcolor{gray}{(ACL 2025)}
& 431 MB
& 26.77
& 44.96
& 23.46
& 31.73 \\

& PM-KVQ \textcolor{gray}{(ICLR 2026)}
& 431 MB
& 26.26
& 47.76
& 23.50
& 32.51 \\

& KVmix \textcolor{gray}{(AAAI 2026)}
& 431 MB
& 26.26
& 45.11
& 25.55
& 32.31 \\

\cmidrule[0.35pt]{2-7}
\rowcolor{gray!8}
\cellcolor{white}
& \textbf{MultiAlloc (Ours)}
& \textbf{431 MB}
& \textbf{27.27}
& \textbf{48.67}
& \textbf{23.94}
& \textbf{33.29} \\

\midrule

\multirow{5}{*}{DeepSeek-Qwen-7B}

& FP16
& 448 MB
& 30.81
& 90.37
& 27.48
& 49.55 \\

& OTT \textcolor{gray}{(ACL 2025)}
& 180 MB
& 28.28
& 87.57
& 17.82
& 44.56 \\

& PM-KVQ \textcolor{gray}{(ICLR 2026)}
& 180 MB
& 29.80
& 89.69
& 25.68
& 48.39 \\

& KVmix \textcolor{gray}{(AAAI 2026)}
& 180 MB
& \textbf{31.31}
& 82.11
& 20.28
& 44.57 \\

\cmidrule[0.35pt]{2-7}
\rowcolor{gray!8}
\cellcolor{white}
& \textbf{MultiAlloc (Ours)}
& \textbf{180 MB}
& 30.30
& \textbf{90.14}
& \textbf{26.75}
& \textbf{49.06} \\

\midrule

\multirow{5}{*}{DeepSeek-Qwen-14B}

& FP16
& 1536 MB
& 37.37
& 89.39
& 27.86
& 51.54 \\

& OTT \textcolor{gray}{(ACL 2025)}
& 620 MB
& 35.35
& 74.30
& 24.28
& 44.64 \\

& PM-KVQ \textcolor{gray}{(ICLR 2026)}
& 620 MB
& \textbf{37.37}
& 88.32
& 25.63
& 50.44 \\

& KVmix \textcolor{gray}{(AAAI 2026)}
& 620 MB
& 34.85
& 88.32
& 26.15
& 49.77 \\

\cmidrule[0.35pt]{2-7}
\rowcolor{gray!8}
\cellcolor{white}
& \textbf{MultiAlloc (Ours)}
& \textbf{620 MB}
& \textbf{37.37}
& \textbf{89.01}
& \textbf{26.54}
& \textbf{50.97} \\

\midrule

\multirow{5}{*}{QwQ-32B}

& FP16
& 2048 MB
& 39.39
& 95.53
& 28.03
& 54.32 \\

& OTT \textcolor{gray}{(ACL 2025)}
& 820 MB
& 37.88
& 95.22
& 26.57
& 53.22 \\

& PM-KVQ \textcolor{gray}{(ICLR 2026)}
& 820 MB
& 37.88
& 95.07
& 27.82
& 53.59 \\

& KVmix \textcolor{gray}{(AAAI 2026)}
& 820 MB
& \textbf{38.38}
& 93.03
& 27.73
& 53.05 \\

\cmidrule[0.35pt]{2-7}
\rowcolor{gray!8}
\cellcolor{white}
& \textbf{MultiAlloc (Ours)}
& \textbf{820 MB}
& \textbf{38.38}
& \textbf{95.38}
& \textbf{28.15}
& \textbf{53.97} \\

\bottomrule
\end{tabular}

\caption{
Main performance comparison of compressed methods under the same
KV cache memory budgets.
}
\label{tab:main_results}
\end{table*}

This advantage remains evident as model scale increases. On
DeepSeek-Qwen-14B, \textit{MultiAlloc} matches the best compressed GPQA result
and achieves the strongest performance on GSM8K and LongBench-v2.
On QwQ-32B, it achieves the best GPQA and LongBench-v2 performance under the
same budget. The smaller gains over existing methods on larger models arise
because their compressed performance already approaches FP16, leaving limited
headroom for improvement. Nevertheless, \textit{MultiAlloc} closes
48.2\% and 52.1\% of the gaps on the 14B and 32B models,
respectively.

Although individual precision preservation strategies may perform well on
specific benchmarks due to task-dependent sensitivity, \textit{MultiAlloc}
provides more balanced performance across reasoning and long-context tasks by
adaptively allocating quantization resources according to heterogeneous
precision demands.

\subsection{Ablation Studies}
\label{sec:ablation}
We conduct ablation studies on Llama-3-8B-Instruct under a fixed 423~MB KV cache
budget. All variants follow the same evaluation settings as the main
experiments. We perform two analyses: precision dimension ablation for layer-
wise precision enhancement and token-wise preservation, and allocation
ablation for profiling-guided precision allocation against manual budget
splits.

Table~\ref{tab:component_ablation_llama3} reports the precision preservation
ablation results.
\textit{Layer-only} removes token preservation, while \textit{Token-only}
removes layer-wise precision enhancement and preserves selected tokens under a
uniform low-bit configuration.
\textit{MultiAlloc} allocates precision based on both layer-wise and token-wise
precision demands.
Removing either dimension degrades performance, confirming their complementary
roles.

\begin{table}[t]
\centering
\small
\setlength{\tabcolsep}{7pt}
\renewcommand{\arraystretch}{1.12}

\begin{tabular}{lccc}
\toprule
Variant
& \makecell{Layer\\alloc.}
& \makecell{Token\\presv.}
& LongBench-v2 \\
\midrule

Layer-only
& $\checkmark$
& $\times$
& 25.58 \\

Token-only
& $\times$
& $\checkmark$
& 24.88 \\

\cmidrule[0.35pt](lr){1-4}

\textbf{MultiAlloc (Ours)}
& $\checkmark$
& $\checkmark$
& \textbf{25.85} \\

\bottomrule
\end{tabular}

\caption{
Component ablation of layer-wise allocation and token-wise preservation.
}
\label{tab:component_ablation_llama3}
\end{table}

Table~\ref{tab:budget_allocation_ablation_llama3} compares the
profiling-guided allocation of \textit{MultiAlloc} with manually specified
allocation strategies between layer-wise precision enhancement and token
preservation.
The manual policies adopt 50/50, 65/35, and 35/65 budget splits, while
\textit{MultiAlloc} determines the allocation through offline profiling.
The 65/35 split achieves comparable performance because it happens to closely
match the allocation selected by \textit{MultiAlloc} under this specific
budget and model setting.

\begin{table}[t]
\centering
\small
\setlength{\tabcolsep}{9pt}
\renewcommand{\arraystretch}{1.12}

\begin{tabular}{lcc}
\toprule
Allocation strategy
& GSM8K
& LongBench-v2 \\
\midrule

Manual Split (65/35)
& 80.29
& 25.59 \\

Manual Split (50/50)
& 77.79
& 25.83 \\

Manual Split (35/65)
& 77.41
& 25.56 \\

\cmidrule[0.35pt](lr){1-3}

\textbf{MultiAlloc (Profiled)}
& \textbf{80.36}
& \textbf{25.85} \\

\bottomrule
\end{tabular}

\caption{
Comparison between profiling-guided budget allocation and manual budget splits.
}
\label{tab:budget_allocation_ablation_llama3}
\end{table}

\subsection{Different Memory Budgets Performance}
\label{sec:budget_results}

We evaluate \textit{MultiAlloc} under multiple KV cache budgets to assess its
robustness under varying memory constraints.
Experiments are conducted on Llama-3-8B-Instruct, Mistral-7B-Instruct, and
DeepSeek-Qwen-14B using GSM8K.
For each model, we select four model-specific budget points and compare OTT,
PM-KVQ, KVmix, and \textit{MultiAlloc} under identical prompting, decoding,
and evaluation settings.

\begin{table}[!htbp]
\centering
\small
\setlength{\tabcolsep}{6pt}
\renewcommand{\arraystretch}{1.12}

\begin{tabular}{lcccc}
\toprule
Method / Budget
& $B_1$
& $B_2$
& $B_3$
& $B_4$ \\
\midrule

\multicolumn{5}{l}{
\textbf{Llama-3-8B-Instruct}
\quad \textit{(FP16: 80.89)}
} \\

Budget (MB)
& 283
& 383
& 423
& 463 \\

\cmidrule(lr){1-5}

OTT
& \underline{76.88}
& 77.56
& 76.95
& 78.09 \\

PM-KVQ
& 58.83
& \underline{79.00}
& \underline{79.98}
& \underline{78.62} \\

KVmix
& 75.13
& 74.07
& 74.83
& 75.21 \\

\cmidrule[0.35pt](lr){1-5}

\textbf{MultiAlloc (Ours)}
& \textbf{77.33}
& \textbf{79.15}
& \textbf{80.36}
& \textbf{80.29} \\

\midrule

\multicolumn{5}{l}{
\textbf{Mistral-7B-Instruct}
\quad \textit{(FP16: 48.75)}
} \\

Budget (MB)
& 311
& 351
& 391
& 431 \\

\cmidrule(lr){1-5}

OTT
& 43.52
& 44.88
& 45.64
& 44.96 \\

PM-KVQ
& \underline{47.69}
& \underline{49.20}
& \textbf{47.31}
& \underline{47.76} \\

KVmix
& 41.62
& 43.52
& 44.12
& 45.11 \\

\cmidrule[0.35pt](lr){1-5}

\textbf{MultiAlloc (Ours)}
& \textbf{48.22}
& \textbf{49.36}
& \underline{47.01}
& \textbf{48.67} \\

\midrule

\multicolumn{5}{l}{
\textbf{DeepSeek-Qwen-14B}
\quad \textit{(FP16: 89.39)}
} \\

Budget (MB)
& 500
& 540
& 580
& 620 \\

\cmidrule(lr){1-5}

OTT
& 72.86
& 72.48
& 73.92
& 74.30 \\

PM-KVQ
& \underline{87.49}
& \underline{87.95}
& \underline{88.10}
& \underline{88.32} \\

KVmix
& 85.97
& 86.13
& 87.57
& \underline{88.32} \\

\cmidrule[0.35pt](lr){1-5}

\textbf{MultiAlloc (Ours)}
& \textbf{89.23}
& \textbf{88.25}
& \textbf{88.70}
& \textbf{89.01} \\

\bottomrule
\end{tabular}

\caption{
Comparison of GSM8K accuracy (\%) across multiple models under different KV cache budgets.
}
\label{tab:budget_comparison}
\vspace{-10pt}
\end{table}

As shown in Table~\ref{tab:budget_comparison}, \textit{MultiAlloc} consistently
achieves strong performance across different models and KV cache budgets.
Under tighter memory constraints, it better preserves inference accuracy than
existing precision preservation strategies, while maintaining stable gains as
the available budget increases.

These results demonstrate that \textit{MultiAlloc} generalizes across different
memory regimes by adaptively allocating limited quantization resources according
to profiled precision demands, achieving a robust accuracy-memory trade-off
under varying KV cache budgets.

\subsection{Analysis of Allocation Behavior}
\label{sec:allocation_analysis}

\begin{figure}[t]
    \centering
    
    \begin{subfigure}{\columnwidth}
        \centering
        \includegraphics[width=\columnwidth]{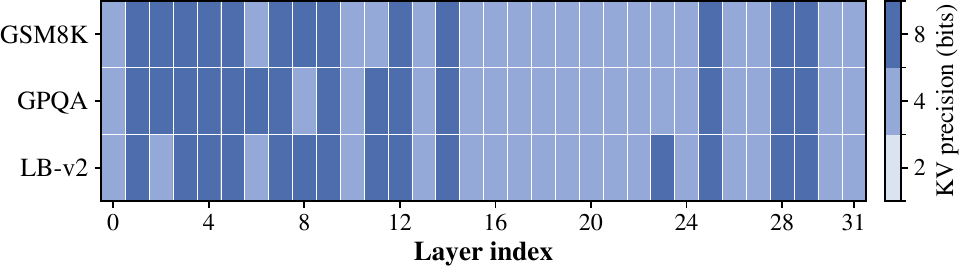}
        \caption{
            Layer-wise KV cache bit-width allocation.
        }
        \label{fig:layer_allocation}
    \end{subfigure}

    \begin{subfigure}{\columnwidth}
        \centering
        \includegraphics[width=\columnwidth]{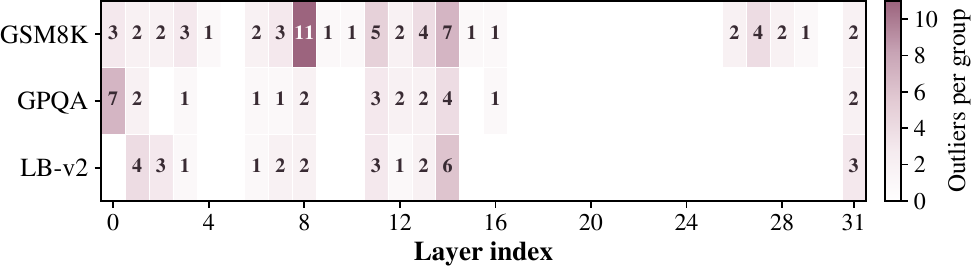}
        \caption{
            Token preservation capacity allocation.
        }
        \label{fig:token_allocation}
    \end{subfigure}

    \caption{
    Allocation behavior of \textit{MultiAlloc} across different tasks under heterogeneous precision demands.
}
    \label{fig:allocation_behavior}
\end{figure}

We further analyze the allocation policies produced by \textit{MultiAlloc} on
Llama-3-8B-Instruct across GSM8K, GPQA, and LongBench-v2.
As shown in Figure~\ref{fig:allocation_behavior}, \textit{MultiAlloc}
generates non-uniform allocation policies with both shared structures and
task-dependent variations, reflecting heterogeneous precision demands across
workloads.

Figure~\ref{fig:allocation_behavior}(a) presents the layer-wise KV cache
bit-width allocation produced by \textit{MultiAlloc}.
The allocation exhibits clear layer-wise variation, where sensitive layers
receive higher precision while most layers are assigned lower bit-widths
under the fixed KV cache budget.
Task-dependent differences further show that offline profiling adapts
bit-width allocation according to workload characteristics.

Figure~\ref{fig:allocation_behavior}(b) presents the token preservation
capacity allocation produced by \textit{MultiAlloc}.
Compared with layer-wise bit-width allocation, token preservation exhibits
stronger task-dependent variation.
GSM8K preserves more tokens across a wider range of layers, while GPQA and
LongBench-v2 concentrate preservation on specific layers.
This sparse allocation shows that \textit{MultiAlloc} selectively assigns
precision resources to important tokens.

\subsection{Runtime Efficiency}
\label{sec:efficiency}

\begin{figure}[!t]
\centering
\includegraphics[width=1\linewidth]{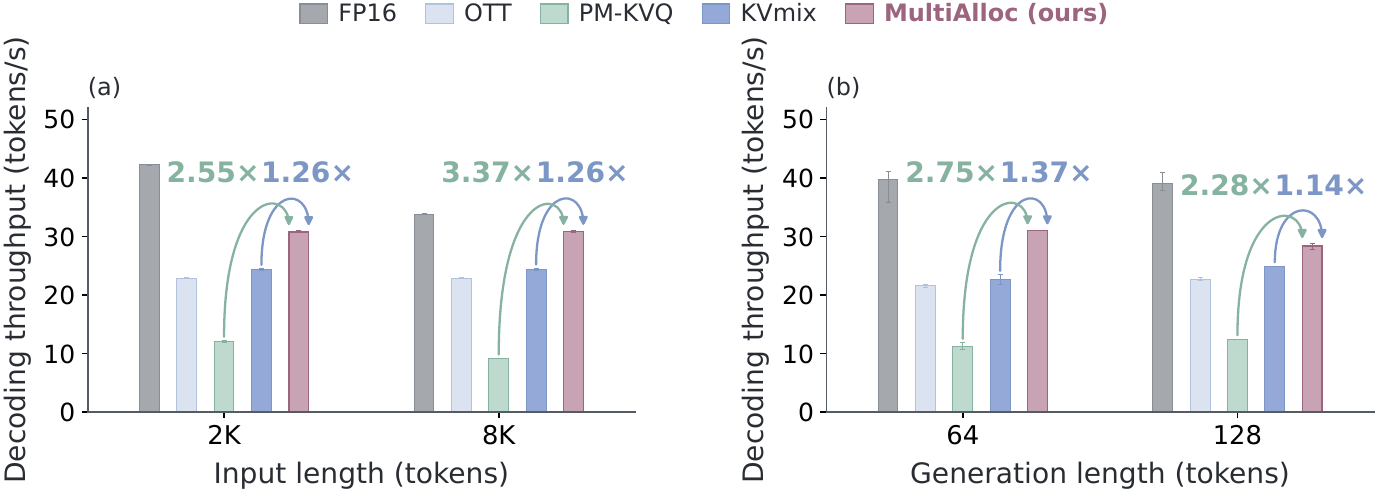}
\caption{
    Runtime efficiency comparison on Llama-3-8B-Instruct under the same
    KV cache budget with varying input and generation lengths.
}
\label{fig:runtime_efficiency}
\end{figure}

We evaluate the runtime efficiency of FP16, OTT, PM-KVQ, KVmix, and
\textit{MultiAlloc} on Llama-3-8B-Instruct using NVIDIA A100 GPUs under the
same 423~MB KV cache budget. We measure decoding throughput under varying
input and generation lengths with batch size one, and average the results over
five runs after warm-up. The one-time offline profiling cost of
\textit{MultiAlloc} is excluded. As shown in Figure~\ref{fig:runtime_efficiency},
\textit{MultiAlloc} consistently achieves higher decoding throughput than
quantized baselines across different input and generation lengths. It
delivers 1.25--1.26$\times$ speedup over the fastest quantized baseline
(KVmix) under varying input lengths and 1.13--1.41$\times$ speedup under
different generation lengths, demonstrating efficient static policy execution
during inference.

\section{Conclusion}

In this work, we formulate KV cache quantization under a fixed KV cache memory
budget as a precision allocation problem, where limited quantization resources
are adaptively distributed between layer-wise precision enhancement and
token-wise preservation.
We propose \textit{MultiAlloc}, a profiling-guided framework that characterizes
heterogeneous precision demands across layers and tokens, and derives a static
precision policy through budget-coupled allocation.
Extensive experiments on reasoning and long-context benchmarks demonstrate that
\textit{MultiAlloc} consistently outperforms representative layer-wise,
token-wise, and importance-based KV cache quantization methods under matched
KV cache memory budgets.

 \section{Limitations}

This work focuses on profiling-guided precision allocation for KV cache
quantization under fixed memory budgets. While \textit{MultiAlloc}
eliminates online allocation search through offline policy derivation,
the profiling stage introduces preprocessing overhead and requires
representative calibration data.
Future work will explore lightweight profiling and improved policy
generalization.

\bibliography{custom}

\begin{thebibliography}{23}
\providecommand{\natexlab}[1]{#1}

\bibitem[{Bai et~al.(2025)Bai, Tu, Zhang, Peng, Wang, Lv, Cao, Xu, Hou, Dong
  et~al.}]{bai2025longbench}
Yushi Bai, Shangqing Tu, Jiajie Zhang, Hao Peng, Xiaozhi Wang, Xin Lv, Shulin
  Cao, Jiazheng Xu, Lei Hou, Yuxiao Dong, et~al. 2025.
\newblock {LongBench v2}: Towards deeper understanding and reasoning on
  realistic long-context multitasks.
\newblock In \emph{Proceedings of the 63rd Annual Meeting of the Association
  for Computational Linguistics (Volume 1: Long Papers)}, pages 3639--3664.

\bibitem[{Cobbe et~al.(2021)Cobbe, Kosaraju, Bavarian, Chen, Jun, Kaiser,
  Plappert, Tworek, Hilton, Nakano et~al.}]{cobbe2021training}
Karl Cobbe, Vineet Kosaraju, Mohammad Bavarian, Mark Chen, Heewoo Jun, Lukasz
  Kaiser, Matthias Plappert, Jerry Tworek, Jacob Hilton, Reiichiro Nakano,
  et~al. 2021.
\newblock Training verifiers to solve math word problems.
\newblock \emph{arXiv preprint arXiv:2110.14168}.

\bibitem[{{DeepSeek-AI}(2025)}]{deepseekai2025deepseekr1incentivizingreasoningcapability}
{DeepSeek-AI}. 2025.
\newblock \href {https://arxiv.org/abs/2501.12948} {{DeepSeek-R1}:
  Incentivizing reasoning capability in {LLMs} via reinforcement learning}.
\newblock \emph{Preprint}, arXiv:2501.12948.

\bibitem[{Grattafiori et~al.(2024)Grattafiori, Dubey, Jauhri, Pandey, Kadian,
  Al-Dahle, Letman, Mathur, Schelten, Vaughan et~al.}]{grattafiori2024llama}
Aaron Grattafiori, Abhimanyu Dubey, Abhinav Jauhri, Abhinav Pandey, Abhishek
  Kadian, Ahmad Al-Dahle, Aiesha Letman, Akhil Mathur, Alan Schelten, Alex
  Vaughan, et~al. 2024.
\newblock The llama 3 herd of models.
\newblock \emph{arXiv preprint arXiv:2407.21783}.

\bibitem[{Ha et~al.(2026)Ha, Lee, and Jeon}]{ha2026kvcobra}
Sihyeon Ha, Jaeho Lee, and Yo-Seb Jeon. 2026.
\newblock \href {https://arxiv.org/abs/2609.24298} {{KV-COBRA}: {KV} cache
  compression via co-optimized bit-rank allocation}.
\newblock \emph{arXiv preprint arXiv:2609.24298}.

\bibitem[{He et~al.(2024)He, Zhang, Wu, Liu, Zhou, and
  Zhuang}]{NEURIPS2024_7e57131f}
Yefei He, Luoming Zhang, Weijia Wu, Jing Liu, Hong Zhou, and Bohan Zhuang.
  2024.
\newblock {ZipCache}: Accurate and efficient kv cache quantization with salient
  token identification.
\newblock In \emph{Advances in Neural Information Processing Systems}.

\bibitem[{Jiang et~al.(2023)Jiang, Sablayrolles, Mensch, Bamford, Chaplot,
  de~las Casas, Bressand, Lengyel, Lample, Saulnier, Renard~Lavaud, Lachaux,
  Stock, Le~Scao, Lavril, Wang, Lacroix, and El~Sayed}]{jiang2023mistral}
Albert~Q. Jiang, Alexandre Sablayrolles, Arthur Mensch, Chris Bamford,
  Devendra~Singh Chaplot, Diego de~las Casas, Florian Bressand, Gianna Lengyel,
  Guillaume Lample, Lucile Saulnier, L{\'e}lio Renard~Lavaud, Marie-Anne
  Lachaux, Pierre Stock, Teven Le~Scao, Thibaut Lavril, Thomas Wang,
  Timoth{\'e}e Lacroix, and William El~Sayed. 2023.
\newblock Mistral 7b.
\newblock \emph{arXiv preprint arXiv:2310.06825}.

\bibitem[{Jiang et~al.(2025)Jiang, Gao, Zarch, and
  Annavaram}]{jiang-etal-2025-kvpr}
Chaoyi Jiang, Lei Gao, Hossein~Entezari Zarch, and Murali Annavaram. 2025.
\newblock {KVPR}: Efficient llm inference with i/o-aware kv cache partial
  recomputation.
\newblock In \emph{Findings of the Association for Computational Linguistics:
  ACL 2025}.

\bibitem[{Kwon et~al.(2023)Kwon, Li, Zhuang, Sheng, Zheng, Yu, Gonzalez, Zhang,
  and Stoica}]{10.1145/3600006.3613165}
Woosuk Kwon, Zhuohan Li, Siyuan Zhuang, Ying Sheng, Lianmin Zheng, Cody~Hao Yu,
  Joseph Gonzalez, Hao Zhang, and Ion Stoica. 2023.
\newblock Efficient memory management for large language model serving with
  {PagedAttention}.
\newblock In \emph{Proceedings of the 29th Symposium on Operating Systems
  Principles}. Association for Computing Machinery.

\bibitem[{Li et~al.(2026)Li, Liu, Wu, Nie, and Wang}]{li2026kvmix}
Fei Li, Song Liu, Weiguo Wu, Shiqiang Nie, and Jinyu Wang. 2026.
\newblock \href {https://doi.org/10.1609/aaai.v40i37.40422} {{KVmix}:
  Gradient-based layer importance-aware mixed-precision quantization for {KV}
  cache}.
\newblock In \emph{Proceedings of the AAAI Conference on Artificial
  Intelligence}, volume~40, pages 31563--31572.

\bibitem[{Li et~al.(2024)Li, Wang, Gu, and Yu}]{li2024xkvpersonalizedkvcache}
Weizhuo Li, Zhigang Wang, Yu~Gu, and Ge~Yu. 2024.
\newblock {XKV}: Personalized kv cache memory reduction for long-context llm
  inference.
\newblock \emph{arXiv preprint arXiv:2412.05896}.

\bibitem[{Li et~al.(2025)Li, XING, Li, Qu, Zhen, Yao, Liu, Pan, and
  Yuan}]{li2025kvtuner}
Xing Li, Zeyu XING, Yiming Li, Linping Qu, Hui-Ling Zhen, Yiwu Yao, Wulong Liu,
  Sinno~Jialin Pan, and Mingxuan Yuan. 2025.
\newblock {KVTuner}: Sensitivity-aware layer-wise mixed-precision kv cache
  quantization for efficient and nearly lossless llm inference.
\newblock In \emph{Forty-second International Conference on Machine Learning}.

\bibitem[{Liu et~al.(2026)Liu, Li, Yang, Zhao, Zhou, Song, Dai, Yan, Yang, and
  Wang}]{liu2026pmkvq}
Tengxuan Liu, Shiyao Li, Jiayi Yang, Tianchen Zhao, Feng Zhou, Xiaohui Song,
  Guohao Dai, Shengen Yan, Huazhong Yang, and Yu~Wang. 2026.
\newblock \href {https://openreview.net/forum?id=Vem6FQvRvq} {{PM-KVQ}:
  Progressive mixed-precision {KV} cache quantization for long-{CoT} {LLM}s}.
\newblock In \emph{The Fourteenth International Conference on Learning
  Representations}.

\bibitem[{Liu et~al.(2024)Liu, Yuan, Jin, Zhong, Xu, Braverman, Chen, and
  Hu}]{liu2024kivi}
Zirui Liu, Jiayi Yuan, Hongye Jin, Shaochen Zhong, Zhaozhuo Xu, Vladimir
  Braverman, Beidi Chen, and Xia Hu. 2024.
\newblock {KIVI}: A tuning-free asymmetric 2bit quantization for kv cache.
\newblock In \emph{Forty-first International Conference on Machine Learning}.

\bibitem[{{Qwen Team}(2025)}]{qwq32b}
{Qwen Team}. 2025.
\newblock \href {https://qwenlm.github.io/blog/qwq-32b/} {Qwq-32b: Embracing
  the power of reinforcement learning}.

\bibitem[{Rein et~al.(2024)Rein, Hou, Stickland, Petty, Pang, Dirani, Michael,
  and Bowman}]{rein2023gpq}
David Rein, Betty~Li Hou, Asa~Cooper Stickland, Jackson Petty, Richard~Yuanzhe
  Pang, Julien Dirani, Julian Michael, and Samuel~R Bowman. 2024.
\newblock {GPQA}: A graduate-level google-proof q\&a benchmark.
\newblock In \emph{First Conference on Language Modeling}.

\bibitem[{Su et~al.(2025)Su, Zhou, Qiu, Li, Xia, Li, Duan, Wang, and
  Zhang}]{su-etal-2025-accurate}
Yi~Su, Yuechi Zhou, Quantong Qiu, Juntao Li, Qingrong Xia, Ping Li, Xinyu Duan,
  Zhefeng Wang, and Min Zhang. 2025.
\newblock Accurate kv cache quantization with outlier tokens tracing.
\newblock In \emph{Proceedings of the 63rd Annual Meeting of the Association
  for Computational Linguistics}.

\bibitem[{Sun et~al.(2026)Sun, He, Harn, and Qin}]{sun2026moend}
Libo Sun, Peixiong He, Po-Wei Harn, and Xiao Qin. 2026.
\newblock \href {https://arxiv.org/abs/2604.17695} {{MoE-nD}: Per-layer
  mixture-of-experts routing for multi-axis {KV} cache compression}.
\newblock \emph{arXiv preprint arXiv:2604.17695}.

\bibitem[{Wang et~al.(2026)Wang, Gu, Li, Wang, Liu, Wang, Li, and
  Yu}]{wang-etal-2026-hqekv}
He~Wang, Yu~Gu, Fangfang Li, Zhigang Wang, Zhenghao Liu, Ning Wang, Xiaohua Li,
  and Ge~Yu. 2026.
\newblock \href {https://doi.org/10.18653/v1/2026.findings-acl.201} {{HqeKV}:
  Towards hybrid quantization and eviction for {KV} cache in long-context {LLM}
  inference}.
\newblock In \emph{Findings of the Association for Computational Linguistics:
  ACL 2026}, pages 4138--4153.

\bibitem[{Wang et~al.(2024)Wang, Cui, and Gan}]{wang2024squeezeattention}
Zihao Wang, Bin Cui, and Shaoduo Gan. 2024.
\newblock {Squeezeattention}: {2D} management of {KV}-cache in {LLM} inference
  via layer-wise optimal budget.
\newblock \emph{arXiv preprint arXiv:2404.04793}.

\bibitem[{Xiao et~al.(2024)Xiao, Tian, Chen, Han, and
  Lewis}]{xiao2024efficient}
Guangxuan Xiao, Yuandong Tian, Beidi Chen, Song Han, and Mike Lewis. 2024.
\newblock Efficient streaming language models with attention sinks.
\newblock In \emph{The Twelfth International Conference on Learning
  Representations}.

\bibitem[{Zhang et~al.(2026)Zhang, Zeng, Peng, Zhuang, and
  Chen}]{zhang-etal-2026-mixkvq}
Tao Zhang, Ziqian Zeng, Hao Peng, Huiping Zhuang, and Cen Chen. 2026.
\newblock \href {https://doi.org/10.18653/v1/2026.acl-long.326} {{MixKVQ}:
  Query-aware mixed-precision {KV} cache quantization for long-context
  reasoning}.
\newblock In \emph{Proceedings of the 64th Annual Meeting of the Association
  for Computational Linguistics (Volume 1: Long Papers)}, pages 7189--7204.

\bibitem[{Zhang et~al.(2023)Zhang, Sheng, Zhou, Chen, Zheng, Cai, Song, Tian,
  R{\'e}, Barrett, Wang, and Chen}]{NEURIPS2023_6ceefa7b}
Zhenyu Zhang, Ying Sheng, Tianyi Zhou, Tianlong Chen, Lianmin Zheng, Ruisi Cai,
  Zhao Song, Yuandong Tian, Christopher R{\'e}, Clark Barrett,
  Zhangyang~``Atlas'' Wang, and Beidi Chen. 2023.
\newblock {H2O}: Heavy-hitter oracle for efficient generative inference of
  large language models.
\newblock In \emph{Advances in Neural Information Processing Systems}.

\end{thebibliography}

\clearpage
\providecommand{\methodabbr}{\textit{MultiAlloc}}
\providecommand{\TOKLOSS}{\operatorname{\textsc{TokLoss}}}
\providecommand{\TOKMEM}{\operatorname{\textsc{TokMem}}}
\providecommand{\LAYERALLOC}{\operatorname{\textsc{LayerAlloc}}}

\appendix

\section{Experimental Setup and Calibration Details}
\label{app:experimental_setup}

This appendix provides additional details on the evaluated
models and benchmarks, the evaluation protocol, and the
task-specific calibration procedure used by MultiAlloc.
Unless otherwise specified, all compressed methods follow the
same evaluation settings as the FP16 baseline.

\subsection{Models and Benchmarks}
\label{app:models_benchmarks}

\paragraph{Models.}
We evaluate MultiAlloc on five instruction-tuned language
models covering different model families and parameter scales:
Llama-3-8B-Instruct, Mistral-7B-Instruct-v0.3,
DeepSeek-R1-Distill-Qwen-7B,
DeepSeek-R1-Distill-Qwen-14B, and QwQ-32B.
These models range from 7B to 32B parameters and include both
general-purpose instruction-tuned models and reasoning-oriented
models.

\paragraph{GPQA.}
GPQA is a multiple-choice benchmark designed to evaluate
expert-level scientific reasoning. We use the GPQA-Diamond
subset, which contains 198 four-option questions.
Accuracy is reported over the evaluated questions.

\paragraph{GSM8K.}
GSM8K evaluates multi-step mathematical reasoning using
grade-school word problems. The dataset contains 7,473
training examples and 1,319 test examples.
We report accuracy on the test set based on the extracted
final numerical answer.

\paragraph{LongBench-v2.}
LongBench-v2 evaluates long-context understanding and
reasoning in realistic scenarios. It contains 503
multiple-choice questions spanning six task categories:
single-document question answering, multi-document question
answering, long in-context learning, code repository
understanding, long dialogue history understanding, and
long structured-data understanding.
Following the main paper, we report the unweighted
macro-averaged accuracy across the six task categories.

\subsection{Evaluation Protocol}
\label{app:evaluation_protocol}

For each model, the FP16 baseline and all compressed methods
use identical evaluation examples, prompt templates,
tokenization, stopping criteria, answer-scoring procedures,
and greedy decoding. Therefore, the compared methods differ
only in their KV-cache compression strategies while following
the same benchmark-specific evaluation protocol.

\paragraph{GPQA.}
We evaluate GPQA-Diamond on all 198 questions using the
zero-shot direct-choice protocol with option permutation,
following \texttt{lm-evaluation-harness}. We use a fixed
random seed of 0 and greedy decoding, with a maximum
generation length of 1,024 tokens. The generated response is
mapped to the corresponding multiple-choice option for
accuracy evaluation.

\paragraph{GSM8K.}
We evaluate all 1,319 examples from the GSM8K test set.
The final numerical answer is extracted from the generated
response and evaluated using exact match. Generation is
performed with greedy decoding and is capped at 1,024 tokens.
FP16 and all compressed methods use the same generation and
answer-extraction procedure.

\paragraph{LongBench-v2.}
We evaluate all 503 examples in LongBench-v2 and report the
unweighted macro-average accuracy over its six task
categories. All methods use greedy decoding. For
Llama-3-8B-Instruct and Mistral-7B-Instruct-v0.3, the maximum
generation length is 128 tokens. For
DeepSeek-R1-Distill-Qwen-7B,
DeepSeek-R1-Distill-Qwen-14B, and QwQ-32B, the maximum
generation length is 64 tokens, with explicit thinking
disabled. QwQ-32B additionally uses constrained-choice
decoding. Invalid, truncated, or failed generations are
counted as incorrect.

\subsection{Task-Specific Calibration Protocol}
\label{app:calibration_protocol}

MultiAlloc uses a task-specific calibration set for each
evaluated benchmark. In other words, the profiling statistics
used for GSM8K, GPQA, and LongBench-v2 are collected from
separate calibration sets corresponding to the target
workload. For a given model and task, the calibration data are
used to estimate the layer-wise precision demand and
token-wise preservation demand required by the offline
allocator. The resulting allocation policy is then fixed and
applied throughout inference for that workload.

\paragraph{Why task-specific calibration?}
MultiAlloc adopts workload-aware offline profiling because the
precision demand of KV cache quantization is inherently dependent on
the workload characteristics. Rather than deriving a single universal
allocation policy, MultiAlloc estimates the precision requirements
under the target inference workload, where different tasks may induce
distinct activation distributions and token importance patterns. These
variations affect both the relative quantization sensitivity of
Transformer layers and the benefit of preserving selected tokens in
higher precision.

This design is analogous to the calibration process commonly used in
post-training quantization, where representative samples are used to
characterize model behavior before deployment. The calibration data
are only used to collect offline profiling statistics, including
layer-wise quantization sensitivity and token-level preservation
demand, rather than to optimize task-specific model parameters or
access inference labels. As shown in the allocation analysis in the
main paper, different workloads produce a stable layer-precision
backbone with task-dependent variations in both layer-wise precision
assignment and token-wise high-precision preservation.

Therefore, workload-specific calibration enables MultiAlloc to
allocate the limited KV cache budget according to the precision demands
of the target deployment scenario instead of relying on a manually
designed fixed allocation strategy. Importantly, all profiling and
allocation decisions are completed offline. Once the precision policy
is generated, inference directly applies the static policy without
online profiling, search, or dynamic reallocation.

\subsection{Implementation Details}
\label{app:implementation_details}

All experiments are conducted on NVIDIA A100 GPUs with
80\,GB of memory. Our implementation uses Python~3.10,
PyTorch~2.5.1, Transformers~4.49.0, and
Datasets~3.3.2. All methods use the same model checkpoints
and software environment, with a batch size of one.

For each benchmark, FP16 and all compressed methods share
identical preprocessing, prompting, tokenization, decoding,
and stopping criteria. The KV-cache budgets used for the main
comparison are reported in Table~1 of the main paper.

For MultiAlloc, profiling and budget-aware allocation are
performed offline. Each model--dataset pair is profiled for
the corresponding target KV-cache budget, and the resulting
allocation policy is saved and directly loaded during
inference. Consequently, online inference involves no
profiling, candidate-policy enumeration, dynamic programming,
budget search, or adaptive reallocation.

Runtime measurements exclude the one-time offline profiling
cost and are averaged over five runs after warm-up. For
reproducibility, each run records the complete experimental
configuration, software environment, GPU information, and
source-code fingerprint. Model and dataset versions, memory
budgets, decoding settings, and evaluation protocols are
specified in the corresponding experiment configuration
files.

\section{Profiling-based Precision Demand Estimation}
\label{app:profiling}

Directly evaluating the downstream performance of every
candidate precision configuration would require repeated
end-to-end inference and is prohibitively expensive.
Instead, MultiAlloc constructs lightweight profiling
surrogates to characterize the relative precision demand at
different granularities. These surrogates are not intended to
predict the absolute downstream accuracy of a compressed
model. Rather, they provide relative preference signals among
candidate precision-preservation choices during offline
allocation.

Specifically, we separately estimate (i) the sensitivity of
different Transformer layers to KV cache quantization under
different bit-widths and (ii) the benefit of selectively
preserving important tokens in higher precision.

\subsection{Layer-wise Precision Demand Estimation}
\label{app:layer_profiling}

The layer-wise precision demand is estimated by quantifying the impact
of KV cache quantization under different candidate bit-widths. The
estimation is based on first-order perturbation analysis, where the
quantization-induced changes of KV representations are used to
approximate their influence on the profiling objective.

Let $\mathcal{L}$ denote the profiling loss, instantiated as the
next-token negative log-likelihood on the calibration set
$\mathcal{D}_{cal}$. For a Transformer layer $l$, quantizing its KV cache
to $b$ bits produces:

\begin{equation}
\bar{K}^{(b)}_l = Q_b(K_l), \quad
\bar{V}^{(b)}_l = Q_b(V_l),
\end{equation}

where $Q_b(\cdot)$ denotes the quantization operator with bit-width
$b$. The corresponding quantization perturbations are defined as:

\begin{equation}
\Delta K_l^{(b)}
=
\bar{K}^{(b)}_l-K_l,
\quad
\Delta V_l^{(b)}
=
\bar{V}^{(b)}_l-V_l .
\end{equation}

Treating the KV representations as intermediate variables in the
computation graph, the loss variation caused by quantization can be
approximated using a first-order Taylor expansion:

\begin{equation}\small
\Delta \mathcal{L}^{(b)}_l
\approx
\left\langle
\frac{\partial \mathcal{L}}{\partial K_l},
\Delta K_l^{(b)}
\right\rangle
+
\left\langle
\frac{\partial \mathcal{L}}{\partial V_l},
\Delta V_l^{(b)}
\right\rangle .
\end{equation}

Directly using the signed perturbation may underestimate the
quantization impact, since positive and negative contributions across
different tensor elements can cancel each other. Therefore, we measure
the magnitude of first-order perturbations with the following
non-negative surrogate:

\begin{equation}\small
\Phi_l^{(b)}
=
\left\|
\Delta K_l^{(b)}
\odot
G_l^K
\right\|_1
+
\left\|
\Delta V_l^{(b)}
\odot
G_l^V
\right\|_1 ,
\end{equation}

where

\begin{equation}\small
G_l^K=\frac{\partial\mathcal{L}}{\partial K_l},
\quad
G_l^V=\frac{\partial\mathcal{L}}{\partial V_l},
\end{equation}

and $\odot$ denotes element-wise multiplication. The surrogate
preserves the magnitude of quantization-induced perturbations while
avoiding cancellation among individual KV elements.

The layer-wise precision demand under candidate bit-width $b$ is
estimated as:

\begin{equation}\small
S_l(b)
=
\mathbb{E}_{(x,y)\sim\mathcal{D}_{cal}}
\left[
\Phi_l^{(b)}
\right].
\end{equation}

Given $N_{cal}$ calibration samples, the expectation is approximated by:

\begin{equation}
\begin{aligned}
\hat{S}_l(b)
&=
\frac{1}{N_{cal}}
\sum_{i=1}^{N_{cal}}
\left(
\left\|
\Delta K_{i,l}^{(b)}
\odot
G_{i,l}^{K}
\right\|_1 \right.\\
&\quad\left.
+
\left\|
\Delta V_{i,l}^{(b)}
\odot
G_{i,l}^{V}
\right\|_1
\right).
\end{aligned}
\end{equation}

For each calibration sample, one forward-backward pass is sufficient to
obtain the full-precision KV representations and their gradients. The
collected KV states are then reused to evaluate different candidate
bit-widths through lightweight quantization and surrogate computation,
without additional gradient computation.

The obtained profile
$\{S_l(b)\}_{l=1}^{L}$
characterizes the relative precision demand of different layers under
different quantization levels. These layer-wise demand estimates are
provided to the budget-coupled allocation stage, where precision
candidates are selected according to the quantified quantization
sensitivity under the target KV cache budget.

\subsection{Token-wise Preservation Demand Estimation}
\label{app:token_profiling}

Besides layer-wise precision variation, KV cache quantization also
exhibits heterogeneous token-level precision requirements. Although most
tokens can be stored using the layer-wise base precision, a small subset
of tokens may introduce larger quantization errors and benefit from
additional high-precision preservation. Therefore, we estimate the
token-wise preservation demand to characterize the benefit of allocating
additional memory to preserved tokens.

For calibration sample $i$, layer $l$, and cached token position $t$,
let $K_{i,l,t}$ denote the corresponding key representation. We define
the token importance score as:

\begin{equation}
s_{i,l,t}=\|K_{i,l,t}\|_1 .
\end{equation}

This metric is motivated by the scaled dot-product attention mechanism.
For a query vector $q_{i,l}$ and a cached key representation
$K_{i,l,t}$, the attention logit satisfies:

\begin{equation}
|
\langle q_{i,l}, K_{i,l,t} \rangle
|
\leq
\|q_{i,l}\|_{\infty}
\|K_{i,l,t}\|_1 .
\end{equation}

Therefore, the magnitude of the key representation provides a
lightweight query-independent proxy for the potential contribution of a
token to attention computation. Unlike exact attention scores, which
depend on individual queries during inference, this proxy can be
collected offline and reused for static policy construction.

For each calibration sample $i$ and layer $l$, the eligible cached
tokens are partitioned into contiguous groups

\begin{equation}
\mathcal{G}_{i,l}
=
\{
\mathcal{G}_{i,l,1},
\ldots,
\mathcal{G}_{i,l,n_{i,l}}
\},
\end{equation}

where each group contains at most $G$ tokens, and the final group may
contain fewer than $G$ tokens if the eligible sequence length is not
divisible by the group size.

For a preservation capacity $k_l$, the high-precision token set within
group $\mathcal{G}_{i,l,g}$ is selected as

\begin{equation}
P_{i,l,g}(k_l)
=
\operatorname{TopK}
\left(
\{s_{i,l,t}:t\in\mathcal{G}_{i,l,g}\},
k_l
\right).
\end{equation}

The selected tokens are preserved in higher precision, while the
remaining tokens are stored using the layer-wise base precision
determined by the allocation stage.

To quantify the information not covered by high-precision preservation,
we define the discarded importance mass of each group as

\begin{equation}
\begin{aligned}
\ell^{\mathrm{tok}}_{i,l,g}(k_l)
&=
\sum_{t\in\mathcal{G}_{i,l,g}} s_{i,l,t}
-
\sum_{t\in P_{i,l,g}(k_l)} s_{i,l,t}
\\
&=
\sum_{t\in
\mathcal{G}_{i,l,g}\setminus P_{i,l,g}(k_l)}
s_{i,l,t}.
\end{aligned}
\end{equation}

This quantity measures the accumulated importance of tokens that are not
preserved in high precision. A smaller value indicates that the selected
preservation capacity covers more influential tokens and therefore
provides stronger protection for important KV representations.

For a complete token-preservation configuration
$\mathbf{k}=\{k_l\}_{l=1}^{L}$, the token-wise preservation demand is
estimated by aggregating the group-level discarded importance over the
calibration traces:

\begin{equation}
L_{\mathrm{tok}}(\mathbf{k})
=
\frac{1}{N_{cal}}
\sum_{i=1}^{N_{cal}}
\frac{1}{L}
\sum_{l=1}^{L}
\frac{1}{n_{i,l}}
\sum_{g=1}^{n_{i,l}}
\ell^{\mathrm{tok}}_{i,l,g}(k_l).
\end{equation}

The obtained token preservation profile characterizes the benefit of
allocating additional high-precision storage across layers. Together
with the layer-wise precision demand profile $\{S_l(b)\}$, it provides a
complementary signal for the budget-coupled allocation stage, where the
limited KV cache budget is distributed between layer-wise precision
enhancement and token-wise preservation.

\subsection{Complementary Precision Demands}
\label{app:complementary_demands}

The layer-wise and token-wise profiling stages characterize different
dimensions of precision heterogeneity in KV cache quantization. Although
both signals reflect the impact of quantization, they capture
complementary precision demands at different granularities.

The layer-wise precision demand describes the quantization sensitivity of
each Transformer layer under different bit-width configurations. It
determines where increasing the base precision of the KV cache can
provide larger benefits. In contrast, the token-wise preservation demand
focuses on the additional benefit of retaining selected tokens in higher
precision within each layer. It identifies where extra memory allocation
can better preserve important token representations.

These two precision demands correspond to different allocation choices.
Layer-wise precision enhancement provides broad protection by improving
the base representation quality of sensitive layers, while token-wise
preservation provides selective protection for a sparse subset of tokens
that remain vulnerable under the assigned layer precision.

Under a fixed KV cache memory budget, these two strategies compete for
the same limited quantization resources. Allocating more memory to
increase layer precision reduces the available budget for token
preservation, while preserving more high-precision tokens limits the
memory available for layer-wise enhancement. Therefore, optimizing either
precision dimension independently cannot achieve the best allocation
under a constrained memory budget.

The two profiling signals are consequently combined in the subsequent
budget-coupled allocation stage, where MultiAlloc determines how the
available KV cache budget should be distributed between layer-wise
precision enhancement and token-wise preservation. 

\section{Budget-Coupled Allocation Optimization}

\subsection{Memory Accounting under Fixed KV Cache Budget}

MultiAlloc formulates KV cache quantization as a precision allocation
problem under a fixed memory constraint. Unlike approaches that optimize
layer-wise precision assignment or token preservation separately, the
two precision preservation strategies share the same KV cache budget and
must be considered within a unified allocation framework.

Given a total KV cache budget $B$, a precision allocation policy is
defined by a layer-wise bit-width configuration

\begin{equation}
q=\{q_l\}_{l=1}^{L},
\end{equation}

and a token preservation configuration

\begin{equation}
k=\{k_l\}_{l=1}^{L},
\end{equation}

where $q_l$ denotes the base KV precision assigned to layer $l$, and
$k_l$ represents the number of high-precision tokens preserved within
each token group of layer $l$.

The allocation objective is formulated as minimizing the quantization
degradation under the memory constraint:

\begin{equation}
\min_{q,k} \mathcal{L}(q,k)
\quad
\text{s.t.}
\quad
M(q,k)\leq B .
\end{equation}

The memory consumption of a policy is decomposed into two components:

\begin{equation}
M(q,k)
=
M_{\mathrm{layer}}(q)
+
M_{\mathrm{token}}(k),
\end{equation}

where $M_{\mathrm{layer}}(q)$ denotes the memory consumption of the
layer-wise quantized KV cache, and $M_{\mathrm{token}}(k)$ represents
the additional memory required for preserving selected tokens in higher
precision.

For a given token preservation configuration $k$, the remaining budget
available for layer-wise precision allocation is:

\begin{equation}
B_{\mathrm{layer}}
=
B-M_{\mathrm{token}}(k).
\end{equation}

Therefore, increasing token preservation capacity inevitably reduces
the memory available for layer-wise precision enhancement, while
allocating higher layer precision limits the budget for additional token
preservation. This budget coupling motivates the subsequent allocation
procedure, which determines how the limited KV cache memory should be
distributed between the two precision dimensions.

\subsection{Layer-wise Allocation Solver}

Given a remaining layer-wise budget $B_{\mathrm{layer}}$, the layer-wise
allocation stage selects the precision configuration according to the
profiled layer-wise precision demands.

Specifically, the layer-wise allocation problem is formulated as:

\begin{equation}
q^{*}
=
\arg\min_q
\sum_{l=1}^{L}S_l(q_l)
\end{equation}

subject to:

\begin{equation}
\sum_{l=1}^{L}
M_l(q_l)
\leq
B_{\mathrm{layer}},
\end{equation}

where $S_l(q_l)$ denotes the estimated precision demand of layer $l$
under the assigned bit-width $q_l$.

The objective selects one precision candidate for each layer while
satisfying the given memory constraint. Since the candidate bit-width
set is small, the problem can be efficiently solved using dynamic
programming over a discretized memory grid.

Let $\hat{B}$ denote the discretized budget capacity. The dynamic
programming state represents the minimum accumulated precision demand
after assigning bit-width candidates to the first $l$ layers:

\begin{equation}\small
DP[l,c]
=
\min_b
\{
DP[l-1,c-C_l(b)]
+
S_l(b)
\}.
\end{equation}

After processing all layers, backtracking recovers the selected
layer-wise precision configuration:

\begin{equation}
q^*=\{q_l^*\}_{l=1}^{L}.
\end{equation}

The recovered configuration satisfies the original byte-level memory
constraint because candidate memory costs are conservatively discretized
during optimization.

\subsection{Collective Budget Allocation}

The final allocation stage determines how the fixed KV cache budget
should be distributed between token preservation and layer-wise
precision enhancement. Instead of manually specifying the budget ratio
between the two strategies, MultiAlloc evaluates different
cross-granularity allocation candidates under the same memory constraint.

Specifically, MultiAlloc constructs a finite set of token preservation
configurations:

\begin{equation}
\mathcal{K}=\{k_i\}_{i=1}^{N},
\end{equation}

where each candidate $k_i$ specifies the layer-wise token preservation
capacities.

For each candidate token configuration $k_i$, its memory consumption is
first calculated as:

\begin{equation}
B_{\mathrm{token}}(k_i)
=
M_{\mathrm{token}}(k_i).
\end{equation}

Candidates whose token preservation memory exceeds the total KV cache
budget are discarded. For each feasible candidate, the remaining budget
for layer-wise precision allocation is computed as:

\begin{equation}
B_{\mathrm{layer}}(k_i)
=
B-B_{\mathrm{token}}(k_i).
\end{equation}

Given the remaining budget, the layer-wise allocation solver introduced
in Section C.2 obtains the corresponding precision configuration:

\begin{equation}
q^*(k_i)
=
\arg\min_q
\sum_{l=1}^{L}S_l(q_l),
\end{equation}

subject to the layer-wise memory constraint.

Each complete allocation candidate is evaluated using the collective
surrogate objective:

\begin{equation}
J(k_i)
=
L_{\mathrm{tok}}(k_i)
+
\sum_{l=1}^{L}
S_l(q_l^*(k_i)).
\end{equation}

The selected token preservation configuration is:

\begin{equation}
k^*
=
\arg\min_{k_i\in\mathcal{K}}
J(k_i),
\end{equation}

which yields the final static precision policy:

\begin{equation}
\pi^*
=
(q^*,k^*).
\end{equation}

By jointly evaluating layer-wise precision enhancement and token-wise
preservation under the shared KV cache budget, MultiAlloc determines the
cross-granularity allocation before inference. The resulting policy is
fixed during deployment and requires no online allocation search.

\section{Algorithm and Static Deployment Details}

\subsection{Offline Profiling Procedure}

The offline profiling stage collects the precision demand statistics
required for subsequent budget-coupled allocation. Given a pretrained
model $f$ and calibration set $\mathcal{D}_{cal}$, the procedure first
obtains full-precision KV representations and their gradients through
forward-backward passes. These intermediate states are then reused to
evaluate different quantization candidates without additional gradient
computation.

For each layer $l$ and candidate bit-width $b$, the quantization
perturbations are used to compute the layer-wise precision demand profile
$\{S_l(b)\}_{l=1}^{L}$. Meanwhile, token importance scores
$s_{l,t}=\|K_{l,t}\|_1$ are collected as the token-level profiling trace,
which is later used to evaluate different token preservation
configurations.

The complete offline profiling procedure is summarized in
Algorithm~\ref{alg:profiling}. It outputs the layer-wise precision demand
table and token-level profiling trace, which remain fixed during policy
selection and inference.

\begin{algorithm}
\caption{Offline Profiling for Layer-wise Precision Demand and Token-level Trace}
\label{alg:profiling}
\begin{algorithmic}[1]

\Require
Pretrained model $f$;
calibration set $\mathcal{D}_{cal}$;
candidate bit-widths $\mathcal{B}$;
quantization operators $\{Q_b(\cdot)\}_{b\in\mathcal{B}}$

\Ensure
Layer-wise precision demand table $\{S_l(b)\}$;
token-level profiling trace $\mathcal{T}$

\State Initialize $S_l(b)\leftarrow 0$ for all $l\in\{1,\dots,L\}$ and $b\in\mathcal{B}$
\State Initialize $\mathcal{T}\leftarrow \emptyset$

\For{each calibration sample $(x_i,y_i)\in\mathcal{D}_{cal}$}

    \State Run one forward-backward pass on $(x_i,y_i)$

    \State Collect
    $\{K_{i,l},V_{i,l},G^K_{i,l},G^V_{i,l}\}_{l=1}^{L}$

    \For{$l=1$ to $L$}

        \State Append token scores
        $\{(i,l,t,\|K_{i,l,t}\|_1)\}_{t}$
        to $\mathcal{T}$

        \For{each $b\in\mathcal{B}$}

            \State Compute quantization perturbations:
            \[
            \Delta K^{(b)}_{i,l}=Q_b(K_{i,l})-K_{i,l}
            \]
            \[
            \Delta V^{(b)}_{i,l}=Q_b(V_{i,l})-V_{i,l}
            \]

            \State Update layer demand:
\[
\begin{aligned}
S_l(b)\leftarrow S_l(b)
&+
\|\Delta K^{(b)}_{i,l}\odot G^K_{i,l}\|_1\\
&+
\|\Delta V^{(b)}_{i,l}\odot G^V_{i,l}\|_1
\end{aligned}
\]

        \EndFor

    \EndFor

\EndFor

\For{$l=1$ to $L$}
    \For{each $b\in\mathcal{B}$}
        \State $S_l(b)\leftarrow S_l(b)/|\mathcal{D}_{cal}|$
    \EndFor
\EndFor

\State \Return $\{S_l(b)\}$ and $\mathcal{T}$

\end{algorithmic}
\end{algorithm}

\subsection{Offline Policy Selection}

After obtaining the profiling statistics, the offline allocation stage
constructs the final precision policy under the target KV cache budget.
Given the layer-wise precision demand table and token-level profiling
trace, the procedure evaluates a finite set of token preservation
configurations and determines their corresponding layer-wise precision
assignments.

For each candidate token configuration $k$, its memory consumption is
first evaluated. Candidates that violate the total KV cache budget are
discarded. For each feasible candidate, the remaining memory budget is
assigned to layer-wise precision allocation, and the corresponding
layer-wise precision configuration is obtained using the solver described
in Section C.2.

The complete allocation procedure is summarized in
Algorithm~\ref{alg:allocation}. The selected configuration minimizes the
collective surrogate objective among all feasible candidates and produces
the final static policy $\pi^*=(q^*,k^*)$. Since all candidate evaluation
and budget allocation are completed offline, no allocation search or
budget adjustment is required during inference.

\begin{algorithm}
\caption{Collective Allocation under a Fixed KV Cache Budget}
\label{alg:allocation}
\begin{algorithmic}[1]

\Require
Layer-wise precision demand table $\{S_l(b)\}$;
token-level profiling trace $\mathcal{T}$;
total KV cache budget $B$;
candidate token configurations $\mathcal{K}$

\Ensure
Static precision policy $\pi^*=(q^*,k^*)$

\State Initialize $J^*\leftarrow +\infty$
\State Initialize $q^*\leftarrow \emptyset$; $k^*\leftarrow \emptyset$

\For{each token configuration $k_i\in\mathcal{K}$}

    \State Compute token preservation memory:
    \[
    B_{\mathrm{token}}(k_i)
    \leftarrow
    M_{\mathrm{token}}(k_i)
    \]

    \If{$B_{\mathrm{token}}(k_i)>B$}
        \State Continue
    \EndIf

    \State Compute remaining layer budget:
    \[
    B_{\mathrm{layer}}(k_i)
    \leftarrow
    B-B_{\mathrm{token}}(k_i)
    \]

    \State Obtain layer-wise precision configuration:
\[
\begin{aligned}
q(k_i)
\leftarrow\;&
\textsc{LayerAllocation}
\\
&(\{S_l(b)\},B_{\mathrm{layer}}(k_i))
\end{aligned}
\]

    \If{$q(k_i)=\emptyset$}
        \State Continue
    \EndIf

    \State Compute collective surrogate objective:
    \[
    J(k_i)
    \leftarrow
    L_{\mathrm{tok}}(k_i)
    +
    \sum_{l=1}^{L}S_l(q_l(k_i))
    \]

    \If{$J(k_i)<J^*$}
        \State $J^*\leftarrow J(k_i)$
        \State $k^*\leftarrow k_i$
        \State $q^*\leftarrow q(k_i)$
    \EndIf

\EndFor

\State \Return $\pi^*=(q^*,k^*)$

\end{algorithmic}
\end{algorithm}

\subsection{Static Inference Deployment}

Once the allocation policy $\pi^*$ is obtained, it is directly applied
during inference. For each Transformer layer $l$, the incoming KV states
are quantized using the assigned bit-width $q_l^*$. Meanwhile, the token
preservation configuration $k^*$ determines the number of high-precision
tokens retained within each token group.

Although the identities of preserved tokens depend on the input
sequence, the selection rule itself is fixed by the offline policy.
Specifically, tokens are ranked using the predefined importance metric
and preserved according to the assigned top-$k$ capacity.

During inference, the execution only involves deterministic KV
quantization, token scoring, high-precision token preservation, and
cache maintenance. No profiling, gradient computation, candidate policy
enumeration, budget optimization, or dynamic precision reallocation is
performed online.

By separating allocation decisions from latency-critical inference,
MultiAlloc applies a fixed precision policy during deployment and avoids
any additional allocation search under the target KV cache memory
constraint.

\section{Additional Allocation Analysis}

We further analyze the precision allocation policies generated by
\textit{MultiAlloc} across different model architectures, parameter
scales, and workloads. The goal of this analysis is to understand how
the proposed profiling-guided allocation adapts limited KV cache
resources under different deployment conditions.

\subsection{Layer-wise Precision Allocation Patterns}

\begin{figure*}[t]
    \centering

    \begin{subfigure}[t]{0.49\textwidth}
        \centering
        \includegraphics[width=\linewidth]
        {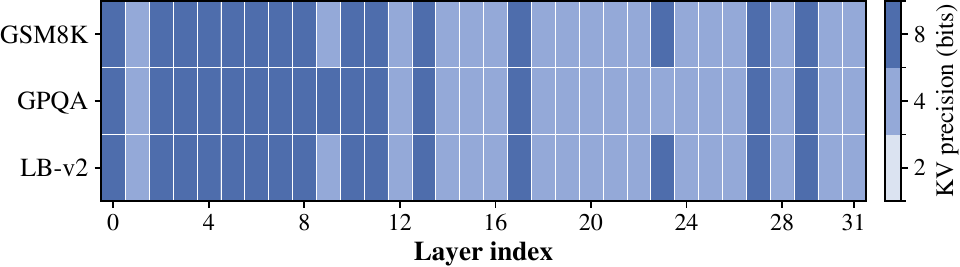}
        \caption{
        Mistral-7B-Instruct-v0.3 under a 431~MB KV cache budget.
        }
        \label{fig:mistral_7b_layer_allocation}
    \end{subfigure}
    \hfill
    \begin{subfigure}[t]{0.49\textwidth}
        \centering
        \includegraphics[width=\linewidth]
        {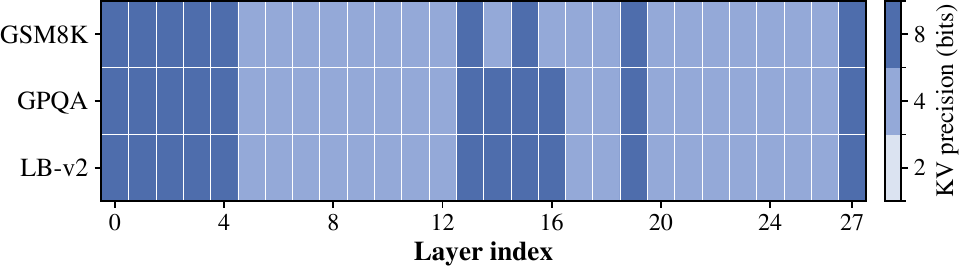}
        \caption{
        DeepSeek-R1-Distill-Qwen-7B under a 180~MB KV cache budget.
        }
        \label{fig:deepseek_qwen_7b_layer_allocation}
    \end{subfigure}

    \caption{
    Layer-wise KV cache precision allocation patterns on the evaluated
    7B models. The assigned bit-widths vary across layers according to
    the profiled precision demands under fixed KV cache budgets.
    }
    \label{fig:cross_model_layer_allocation_7b}
\end{figure*}

\begin{figure*}[t]
    \centering

    \begin{subfigure}[t]{0.49\textwidth}
        \centering
        \includegraphics[width=\linewidth]
        {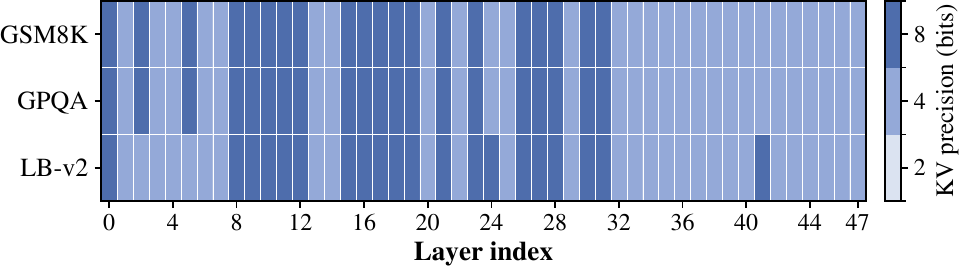}
        \caption{
        DeepSeek-R1-Distill-Qwen-14B under a 620~MB KV cache budget.
        }
        \label{fig:deepseek_qwen_14b_layer_allocation}
    \end{subfigure}
    \hfill
    \begin{subfigure}[t]{0.49\textwidth}
        \centering
        \includegraphics[width=\linewidth]
        {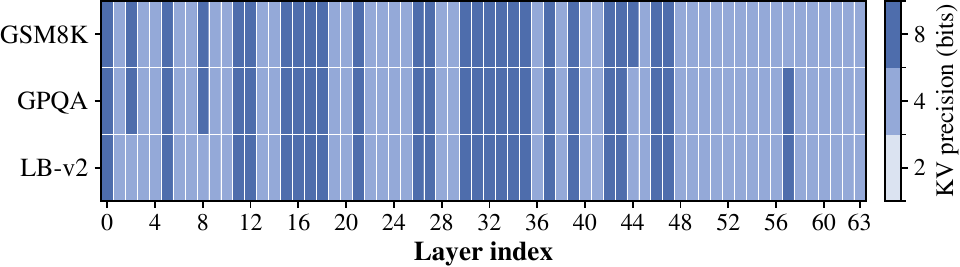}
        \caption{
        QwQ-32B under an 820~MB KV cache budget.
        }
        \label{fig:qwq_32b_layer_allocation}
    \end{subfigure}

    \caption{
    Layer-wise KV cache precision allocation patterns on larger-scale
    models. The distinct distributions demonstrate model-dependent
    precision demands and the necessity of profiling-guided allocation.
    }
    \label{fig:cross_model_layer_allocation_large}
\end{figure*}

Figures~\ref{fig:cross_model_layer_allocation_7b} and
\ref{fig:cross_model_layer_allocation_large} present the layer-wise KV
cache precision configurations under different models and memory budgets.
Each heatmap visualizes the assigned bit-width of individual Transformer
layers derived from the profiling-guided allocation process.

As shown in Figures~\ref{fig:cross_model_layer_allocation_7b} and
\ref{fig:cross_model_layer_allocation_large}, the allocated precision
levels exhibit clear layer-wise variation rather than a uniform
configuration across all Transformer layers. Layers with larger
quantization-induced degradation receive higher precision candidates,
while other layers are assigned lower precision to satisfy the target
memory constraint.

The allocation patterns further vary across model scales and workloads,
indicating that layer-wise precision demand is not determined solely by
layer depth or model architecture. Therefore, fixed or manually designed
layer-wise precision policies cannot consistently capture the
heterogeneous quantization requirements under different deployment
settings.

\subsection{Token Preservation Allocation Patterns}

\begin{figure*}[t]
    \centering

    \begin{subfigure}[t]{0.49\textwidth}
        \centering
        \includegraphics[width=\linewidth]
        {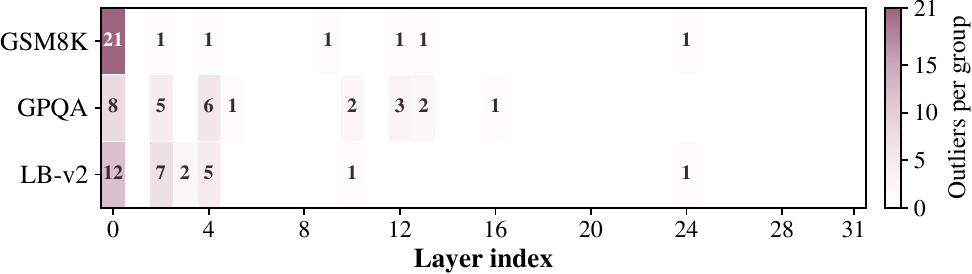}
        \caption{
        Mistral-7B-Instruct-v0.3 under a 431~MB KV cache budget.
        }
        \label{fig:mistral_7b_token_allocation}
    \end{subfigure}
    \hfill
    \begin{subfigure}[t]{0.49\textwidth}
        \centering
        \includegraphics[width=\linewidth]
        {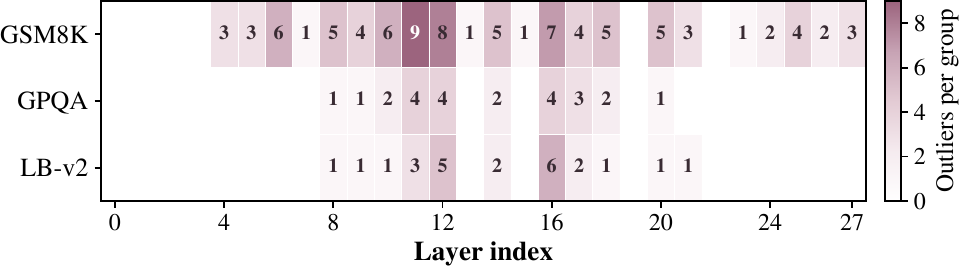}
        \caption{
        DeepSeek-R1-Distill-Qwen-7B under a 180~MB KV cache budget.
        }
        \label{fig:deepseek_qwen_7b_token_allocation}
    \end{subfigure}

    \caption{
    Token preservation capacity allocation patterns on the evaluated 7B
    models. The available preservation budget is distributed
    non-uniformly across layers according to the profiled token-level
    preservation demands.
    }
    \label{fig:cross_model_token_allocation_7b}
\end{figure*}

\begin{figure*}[t]
    \centering

    \begin{subfigure}[t]{0.49\textwidth}
        \centering
        \includegraphics[width=\linewidth]
        {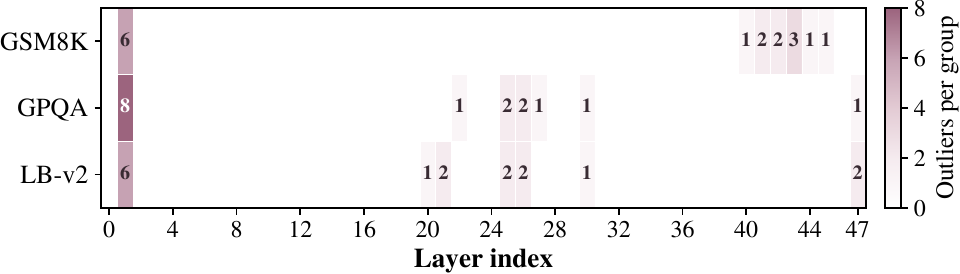}
        \caption{
        DeepSeek-R1-Distill-Qwen-14B under a 620~MB KV cache budget.
        }
        \label{fig:deepseek_qwen_14b_token_allocation}
    \end{subfigure}
    \hfill
    \begin{subfigure}[t]{0.49\textwidth}
        \centering
        \includegraphics[width=\linewidth]
        {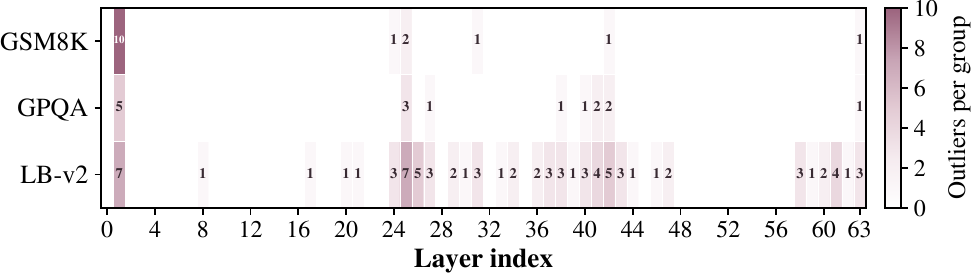}
        \caption{
        QwQ-32B under an 820~MB KV cache budget.
        }
        \label{fig:qwq_32b_token_allocation}
    \end{subfigure}

    \caption{
    Token preservation capacity allocation patterns on larger-scale
    models. The distributions vary across models and tasks, indicating
    workload-dependent token-level preservation demands under fixed KV
    cache budgets.
    }
    \label{fig:cross_model_token_allocation_large}
\end{figure*}

Figures~\ref{fig:cross_model_token_allocation_7b} and
\ref{fig:cross_model_token_allocation_large} analyze how the available
token preservation capacity is distributed across Transformer layers
under different models and workloads.

Unlike layer-wise precision allocation, token preservation introduces an
additional sparse allocation dimension, where only a subset of tokens
receive extra high-precision storage. As shown in
Figures~\ref{fig:cross_model_token_allocation_7b} and
\ref{fig:cross_model_token_allocation_large}, the preservation capacity
is not uniformly distributed across layers. Instead, different layers
receive different preservation capacities according to their estimated
token-level preservation demands.

The allocation patterns further vary across models and workloads,
indicating that the benefit of additional high-precision storage depends
on both model characteristics and task distributions. Therefore, a fixed
preservation ratio or manually predefined token allocation policy cannot
consistently capture the heterogeneous token-level precision demands under
different KV cache constraints.

\subsection{Fixed Split versus Adaptive Allocation}

To further examine the necessity of budget coupling, we compare
\textit{MultiAlloc} with manually specified memory splits between
layer-wise precision enhancement and token preservation.

A fixed split allocates a predefined proportion of the KV cache budget to
each precision preservation strategy, regardless of the underlying model
characteristics, workload, or target memory constraint. However, the
appropriate allocation between the two dimensions may vary across
different deployment settings.

As shown in Table~\ref{tab:budget_allocation_ablation_llama3}, different manual split
ratios result in distinct performance trade-offs. Although a manually
selected ratio can achieve competitive performance in certain settings,
no fixed split consistently provides the best allocation across the
evaluated configurations.

In contrast, \textit{MultiAlloc} determines the budget distribution
between layer-wise precision enhancement and token preservation through
offline profiling and budget-coupled optimization. By avoiding
manually predefined budget partitioning, the proposed allocation strategy
better adapts the limited KV cache memory to heterogeneous precision
demands under different deployment constraints.

\section{Complexity Analysis}

We analyze the computational overhead of \textit{MultiAlloc} by
separating the one-time offline allocation cost from the inference-time
execution cost. The allocation process is performed entirely before
deployment, and inference only applies the resulting static precision
policy without additional search.

\subsection{Offline Complexity}

The offline stage consists of precision demand profiling and
budget-coupled policy selection. Let $N_{cal}$ denote the number of
calibration samples, $L$ the number of Transformer layers, $T$ the
profiling sequence length, $D_{KV}$ the KV representation dimension,
$|\mathcal{B}|$ the number of candidate bit-widths, and
$|\mathcal{K}|$ the number of token preservation candidates.

For each calibration sample, one forward-backward pass is performed to
obtain the KV representations and their gradients. The collected
intermediate states are reused to evaluate different quantization
candidates, leading to the profiling complexity:

\begin{equation}
O
\left(
N_{cal}C_{fb}(T)
+
N_{cal}L|\mathcal{B}|TD_{KV}
\right),
\end{equation}

where $C_{fb}(T)$ denotes the cost of one forward-backward pass.

After profiling, the allocation stage evaluates different token
preservation candidates. For each candidate, the remaining memory budget
is allocated to layer-wise precision selection through the solver
described in Section C.2. The overall offline allocation complexity is:

\begin{equation}
O
\left(
|\mathcal{K}|N_{cal}LT
+
L|\mathcal{B}|\hat{B}
\right),
\end{equation}

where $\hat{B}$ denotes the discretized memory capacity used by the
dynamic programming solver.

Both the candidate bit-width set and token preservation candidate set are
constructed offline and remain small in practice. Therefore, the
allocation overhead is incurred only once for each target model,
calibration configuration, and KV cache budget.

\subsection{Inference-Time Complexity}

After offline optimization, inference directly applies the precomputed
static policy

\begin{equation}
\pi^*=(q^*,k^*).
\end{equation}

During decoding, each Transformer layer quantizes the incoming KV states
using the assigned bit-width $q_l^*$, while token preservation follows
the fixed capacity configuration specified by $k^*$.

The online execution only performs deterministic cache operations,
including KV cache quantization, token scoring, high-precision token
preservation according to the fixed policy, and cache maintenance. No
profiling, candidate enumeration, dynamic programming, budget search, or
precision reallocation is performed during inference.

When the token group size and preservation capacity are bounded, the
additional computation introduced by the allocation policy is
$O(LD_{KV})$ per generated token. This cost is independent of the size of
the offline candidate space and the target KV cache budget.

Therefore, \textit{MultiAlloc} separates the allocation overhead from
latency-critical inference by resolving all precision decisions offline,
enabling efficient deployment under fixed KV cache memory constraints.

\end{document}